\documentclass{article}
\usepackage{iclr2027_conference,times}

\usepackage{amsmath,amsfonts,amssymb,bm}
\usepackage{array,booktabs,tabularx,multirow,float}
\usepackage{graphicx}
\usepackage{microtype}
\usepackage{xcolor}
\usepackage{colortbl}
\usepackage{pifont}
\usepackage{tikz}
\usetikzlibrary{arrows.meta,positioning,calc}
\usepackage{hyperref}
\usepackage{url}
\hypersetup{hidelinks}

\newcommand{\method}{\textsc{Sextant}}
\newcommand{\bench}{PanoPed}
\newcommand{\benchS}{PanoPed-S}
\newcommand{\benchR}{PanoPed-R}
\newcommand{\supportR}{PanoPed-R-Support-auto}
\newcommand{\cmark}{\ding{51}}
\newcommand{\xmark}{\ding{55}}
\newcommand{\gain}[1]{\textcolor{green!45!black}{#1}}
\newcommand{\loss}[1]{\textcolor{red!60!black}{#1}}
\newcommand{\ours}{\cellcolor{blue!8}}
\DeclareMathOperator{\wrap}{wrap}
\DeclareMathOperator{\LN}{LN}
\newcommand{\figfile}[3]{\includegraphics[width=\linewidth,height=#2,keepaspectratio]{#1}}

\newcommand{\PanoPedRFrames}{28{,}002}
\newcommand{\PanoPedRLabeledFrames}{16{,}247}
\newcommand{\PanoPedRDurationSeconds}{934.33}

\newcommand{\PanoPedRTracks}{30}
\newcommand{\PanoPedRBoxes}{151{,}461}
\newcommand{\PanoPedRScoreableBoxes}{150{,}115}
\newcommand{\PanoPedRSeamTracks}{10}
\newcommand{\PanoPedRPolarBoxes}{3{,}151}

\newcommand{\PanoPedSReleaseFrames}{108{,}000}
\newcommand{\PanoPedSIdentityTracks}{1{,}486}
\newcommand{\PanoPedSStaticIdentityTracks}{796}
\newcommand{\PanoPedSEgoIdentityTracks}{690}
\newcommand{\PanoPedSWrapTracks}{799}
\newcommand{\PanoPedSWrapInstanceFrames}{40{,}518}

\newcommand{\PanoPedSHDRIs}{13}

\newcommand{\PanoPedSEngine}{Unreal Engine 5.8.2}
\newcommand{\PanoPedSTrainingArchiveGiB}{78.46}
\newcommand{\PanoPedSCompleteArchiveGiB}{153.70}
\newcommand{\HeadParams}{34{,}692}

\title{\centering PanoPed: Beyond Bounding Boxes for\\
Sim-to-Real Panoramic Pedestrian Tracking}
\author{\makebox[\dimexpr\textwidth-2\tabcolsep\relax][c]{\begin{tabular}{c}
Qinfeng Zhu\textsuperscript{1,2}\quad Weiguang Zhao\textsuperscript{2,3}\quad Yunxi Jiang\textsuperscript{4}\\
Anh Nguyen\textsuperscript{2}\quad Lei Fan\textsuperscript{1}\thanks{Corresponding author: \href{mailto:lei.fan@xjtlu.edu.cn}{lei.fan@xjtlu.edu.cn}.}\\[0.35em]
\small \textsuperscript{1}Xi'an Jiaotong-Liverpool University\\
\small \textsuperscript{2}University of Liverpool\quad
\textsuperscript{3}Duke Kunshan University\\
\small \textsuperscript{4}CNRS\\[0.35em]
\small Project: \url{https://zhuqinfeng1999.github.io/PanoPed/}\\
\small Code: \url{https://github.com/zhuqinfeng1999/PanoPed}
\end{tabular}}}
\iclrfinalcopy
\begin{document}
\maketitle
% The ICLR layout and bibliography are retained; the arXiv edition has no
% conference-publication claim in its running head.
\fancyhead{}
\renewcommand{\headrulewidth}{0pt}

\begin{abstract}
Full-sphere panoramic cameras let fixed monitoring systems and mobile robots track people in every direction, but a planar bounding box does not fully describe where a person is on the sphere. We introduce PanoPed, a sim-to-real benchmark for pedestrian tracking on the full sphere. PanoPed-S contains 108,000 frames from fixed, quadruped-mounted, and drone-mounted cameras, with synchronized masks, depth, camera poses, and 3D pedestrian states. PanoPed-R adds 28,002 real frames from fixed cameras, 16,247 of them densely annotated. We find that an ERP rectangle cannot uniquely determine the spherical center and angular extent of the visible person, while the detector's visual query still carries information about them. Inspired by the sextant's use of angular measurements to locate objects, we propose \textsc{Sextant}, a plug-and-play angular localization head with only about 0.035M parameters. It reuses a frozen detector, keeps track identities unchanged, and needs no extra image encoder. \textsc{Sextant} gives the best result in our PanoPed-S test comparison, raising the strongest baseline, MOTIP, from 47.30 to 49.49 HOTA, with gains on all eight test sequences. Without fine-tuning on real data, the same synthetic-trained heads improve MOTIP and HAT by 0.96--1.14 HOTA on real video, and both seeds improve every real sequence. HAT+\textsc{Sextant} scores best among the compared systems that add no localization image encoder.
\end{abstract}

\section{Introduction}
\label{sec:intro}
Fixed cameras and robots that move through crowds see people on all sides, not
only in front. Full-sphere cameras capture this whole scene, usually in the
equirectangular projection (ERP), whose left and right borders are the same
meridian and whose distortion grows toward the poles. A pedestrian can leave
the image on one side and reappear on the other while staying in view, and a
person near the lower pole appears as a wide, curved shape. A moving camera
adds viewpoint changes and platform motion. This setting poses two problems:
suitable data are missing, and a tracker's box output does not determine where
the visible person is on the sphere.

\begin{figure}[t]
\centering
\figfile{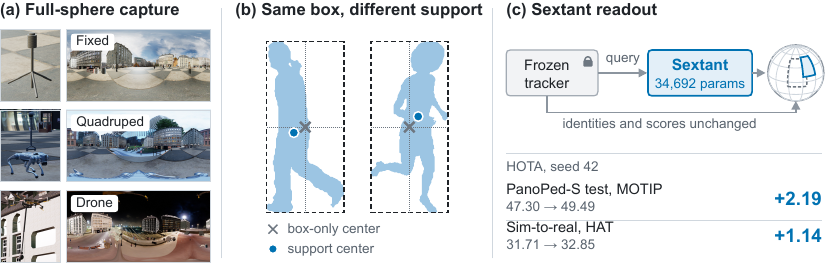}{4.5cm}{}
\caption{\textbf{PanoPed and \method{}.}
(a) Full-sphere observations from fixed, quadruped, and drone cameras.
(b) Conceptual example: the same box contains different visible supports.
Black dashed lines mark the box; gray crosses and blue dots mark box-only
and support-derived centers, respectively.
(c) \method{} refines localization without changing identities. Results
shown use seed 42; both seeds are reported in Section~\ref{sec:experiments}.
Sim-to-real evaluation uses the support targets in Section~\ref{sec:panopedr}.
Silhouettes are two rendered \benchS{} masks rescaled to one box.}
\label{fig:teaser}
\end{figure}

The first problem is data. Existing datasets cover only parts of this setting:
JRDB-PanoTrack omits the poles \citep{le2024jrdbpanotrack}, QuadTrack and
EmboTrack cover the sphere partly or unevenly
\citep{luo2025omnitrack,luo2025omnitrackpp}, and 360VOT tracks a single object
\citep{huang2023vot}. None scores multi-person tracks on the sphere or offers
dense masks, depth, and 3D states across platforms. Such labels are costly on
real video, so simulation is the natural source, but its value must be shown
on real targets built the same way.

The second problem lies in the tracker's output. Trackers report each person
as an ERP box, which records how far the person extends but not which pixels
inside it belong to them. A fixed rule then converts the box into a region on
the sphere, so two people in different poses who share one box receive the
same region, although their visible pixels have different centers on the
sphere (Figure~\ref{fig:teaser}b). Box-based evaluation hides this error,
because it compares boxes with boxes.

We address the first problem with \bench{}, a sim-to-real benchmark for
pedestrian tracking on the full sphere (Figure~\ref{fig:teaser}a). \benchS{}
contains 60 synthetic sequences from 30 fixed, 18 quadruped-mounted, and 12
drone-mounted cameras whose motion follows simulated platform dynamics, and
every frame has synchronized masks, depth, camera poses, and 3D pedestrian
states. \benchR{} contains five real recordings from fixed cameras, three of
them densely annotated over 16,247 frames. Both parts compute targets on the
sphere from visible pixels by the same rule, with automatic masks on real
video, so a model trained in simulation can be tested on real video directly.

We address the second problem with \textbf{\method{}}, which reads three
quantities that the tracker already computes: the detector's visual query for
the person, a starting region from the box, and the detection score. From them
it predicts the center direction and angular extent of the visible person
(Figure~\ref{fig:teaser}c). The tracker stays frozen, no image encoder is
added, and identities and scores pass through unchanged. With only
\HeadParams{} parameters, \method{} improves MOTIP, the strongest baseline in
our \benchS{} comparison, by 2.17--2.19 test HOTA, with gains on all eight
test sequences for both head seeds.

Because both parts share one target rule, we can test whether this correction
transfers. Without fine-tuning, heads trained only on \benchS{} improve MOTIP
and HAT by 0.96--1.14 HOTA on all 16,247 real frames, with gains in HOTA and
IDF1 on every sequence for both seeds and trackers. HAT+\method{} scores
highest among compared systems that add no localization image encoder.

Our contributions are: \textbf{(i)} \bench{}, a sim-to-real benchmark for
full-sphere pedestrian tracking whose targets are defined on the sphere by one
rule in simulation and on real video; \textbf{(ii)} evidence that a tracker's
box output loses localization information, with the detector and identities
held fixed; and \textbf{(iii)} \method{}, a \HeadParams{}-parameter plug-in
that reads this information from detector queries and transfers from
simulation to real video without a localization image encoder.

\section{Related Work}
\label{sec:related}
\paragraph{Panoramic tracking benchmarks.}
MOTChallenge \citep{milan2016mot16,dendorfer2021motchallenge} and DanceTrack
\citep{sun2022dancetrack} define multi-object tracking with planar boxes in
perspective video. Panoramic datasets widen the view but cover only part of
the sphere or the task. JRDB \citep{martinmartin2021jrdb} and JRDB-PanoTrack
\citep{le2024jrdbpanotrack} use a stitched cylindrical panorama without the
poles. QuadTrack \citep{luo2025omnitrack}, released with OmniTrack, is
recorded through an annular lens on a quadruped, and EmboTrack
\citep{luo2025omnitrackpp} adds dual-fisheye video from a wheel-legged robot,
but its vertical coverage is uneven. 360VOT \citep{huang2023vot} covers the
full sphere and scores regions on it, but tracks one object. No existing
resource combines multi-person tracking, full-sphere coverage, spherical
targets, dense masks and depth, and several platforms
(Table~\ref{tab:resources}).

\paragraph{Localization on the sphere.}
Panoramic perception is moving from distortion-aware engineering toward
sphere-native models \citep{zhu2026panoramicsurvey,zhu2026so3uformer}. Work on
spherical detection follows two lines. One adapts the network to ERP
distortion, with spherical convolutions \citep{coors2018spherenet} or
detectors trained on panoramic images \citep{chou2020indoor360}. The other
changes what is predicted. PANDORA \citep{xu2022pandora} uses oriented boxes
on the sphere, and later work makes them easier to match and regress with an
unbiased spherical IoU \citep{dai2022unbiasediou}, Gaussian label
distributions \citep{xu2023gldl}, or a mapping to planar oriented boxes
(Sph2Pob) \citep{liu2023sph2pob}. Both lines train new single-image detectors.
Neither examines trackers, which are usually trained with ERP boxes and output
boxes, or how much spherical localization a tracker loses at its output.

\paragraph{Multi-object tracking.}
Most progress in multi-object tracking concerns association. ByteTrack
\citep{zhang2022bytetrack} keeps low-confidence detections for matching, and
OC-SORT \citep{cao2023ocsort} corrects its motion model with recent
observations. Transformer trackers extend DETR
\citep{carion2020detr,zhu2021deformabledetr} with track queries
\citep{meinhardt2022trackformer,zeng2022motr}. MOTIP \citep{gao2025motip}
predicts identities from a history of identity-tagged observations, and HAT
\citep{gao2025hat} transforms ReID features with track history on top of MASA
pretraining \citep{li2024masa}. All of these trackers report each person as an
image box. On the full sphere, a fixed rule must still convert this box into a
region, and better association does not reduce the error of that conversion.

\paragraph{Synthetic data and sim-to-real transfer.}
Game engines supply pedestrian video with exact labels. JTA and MOTSynth
render boxes, masks, depth, and pose \citep{fabbri2018jta,fabbri2021motsynth}.
SPEAR is a programmable Unreal Engine simulator \citep{roberts2022spear}.
WorldRover renders annotated exploration video \citep{xu2026worldrover},
whereas Sekai2 collects real exploration video \citep{he2026sekai2}. None of them targets pedestrian
tracking on the full sphere. Transfer to real video also requires both domains
to define targets in the same way, since annotation mismatches already affect
models trained across detection datasets \citep{liao2024labels}. Consistent
real targets need masks, which are costly to label by hand. Box-prompted
segmenters such as SAM~2 \citep{ravi2025sam2} and SAM-HQ \citep{ke2023samhq}
produce masks automatically, but these are estimates rather than verified
labels.

\section{The \bench{} Benchmark}
\label{sec:dataset}

\subsection{Tracking on the sphere}
\label{sec:task}
The input is an ERP image, and the output is one region per visible
pedestrian, each with an identity that persists within the sequence. For
pixel-center indices $(x,y)$ in an image of width $W$ and height $H$, the
spherical coordinates are
\begin{equation}
\lambda=2\pi\left(\frac{x+1/2}{W}-\frac12\right),\qquad
\varphi=\pi\left(\frac12-\frac{y+1/2}{H}\right).
\label{eq:erp_sphere}
\end{equation}
The horizontal wrap-around comes from the image format, so a person who
crosses the seam keeps one identity even when a box export splits them into
two pieces. A native spherical region stores a center direction, an angular
extent, and the frame and boundary convention of the rule that produced it,
and native IoU is computed between these regions on the sphere.

\bench{} provides four linked annotation levels. \textbf{L0} holds the source
annotations: rendered instance masks in S and human boxes around the visible
body in R. \textbf{L1} holds regions on the sphere, \textbf{L2} ERP boxes that
may wrap around the seam, and \textbf{L3} standard MOT rows in which both
pieces of a seam-crossing box keep one identity. S regions summarize the
visible pixels. R has two region layers: \emph{Box-fit}, an oriented and
possibly asymmetric region fitted to each human box, and \emph{Support-auto},
which applies the construction of S to automatically estimated masks. Box-fit
keeps the original scored set, and Support-auto treats observations without a
usable mask by fixed rules (Appendix~\ref{app:support_auto}).

\begin{table}[t]
\centering
\caption{\textbf{\bench{} among pedestrian tracking resources.} Attributes
follow the cited descriptions. \emph{Full sphere}: coverage includes both
poles. \emph{Sph.\ target}: the scored region is defined on the sphere.
\emph{Depth/3D}: per-frame depth or 3D state is released. \emph{Multi-rig}:
more than one camera platform.}
\label{tab:resources}

\begin{minipage}{\linewidth}
\raggedright
\footnotesize
$\dagger$ Parent JRDB supplies LiDAR and 3D boxes, not dense panoramic range.\\
$\ddagger$ EmboTrack combines $360^\circ\!\times70^\circ$ PAL video and
Insta360 dual-fisheye video; full vertical coverage is not uniform.\\
$\ast$ Human-box-guided automatic masks in Support-auto, not human mask labels.
\end{minipage}
\smallskip

\setlength{\tabcolsep}{3.5pt}
\footnotesize
\newcommand{\rot}[1]{\rotatebox{90}{#1}}
\begin{tabularx}{\linewidth}{@{}l>{\raggedright\arraybackslash}Xcccccl@{}}
\toprule
Resource & Domain and platform & \rot{Full sphere} & \rot{Sph.\ target} & \rot{Masks} & \rot{Depth/3D} & \rot{Multi-rig} & Task \\
\midrule
MOTChallenge (2D) \citeyearpar{dendorfer2021motchallenge} & Real, pinhole & \xmark & \xmark & \xmark & \xmark & \cmark & MOT \\
MOTSynth \citeyearpar{fabbri2021motsynth} & Synthetic, pinhole & \xmark & \xmark & \cmark & \cmark & \xmark & MOT \\
JRDB-PanoTrack \citeyearpar{le2024jrdbpanotrack} & Real robot, stitched cylinder & \xmark & \xmark & \cmark & \cmark$^{\dagger}$ & \xmark & Panoptic, MOT \\
360VOT \citeyearpar{huang2023vot} & Real, ERP video & \cmark & \cmark & \xmark & \xmark & \xmark & SOT \\
QuadTrack \citeyearpar{luo2025omnitrack} & Real quadruped, annular lens & \xmark & \xmark & \xmark & \xmark & \xmark & MOT \\
EmboTrack \citeyearpar{luo2025omnitrackpp} & Real, PAL + dual-fisheye & mixed$^{\ddagger}$ & \xmark & \xmark & \xmark & \cmark & MOT \\
\midrule
\ours\textbf{\benchS{}} & \ours Synthetic; fixed, quadruped, drone & \ours\cmark & \ours\cmark & \ours\cmark & \ours\cmark & \ours\cmark & \ours MOT \\
\ours\textbf{\benchR{}} & \ours Real, fixed full-sphere & \ours\cmark & \ours\cmark & \ours\cmark$^{\ast}$ & \ours\xmark & \ours\xmark & \ours MOT \\
\bottomrule
\end{tabularx}
\end{table}

\subsection{\benchS{}: synchronized truth from fixed and moving cameras}
\label{sec:panopeds}
\benchS{} contains 30 fixed-camera, 18 quadruped-mounted, and 12
drone-mounted sequences, each 60 seconds at 30 fps, for
\PanoPedSReleaseFrames{} frames and \PanoPedSIdentityTracks{} sequence-local
identities: 56 sequences at $2048\times1024$ and four fixed TEST
sequences at $3840\times1920$. The environment is a reauthored multilevel
city with plazas, stairs, railings, and street furniture, rendered in
\PanoPedSEngine{} under \PanoPedSHDRIs{} HDRI lighting conditions and varied
layouts. Five fixed sites span camera heights of 0.8--3.0\,m with 5--50
pedestrians and moving-camera sequences use 16--32, drawn from 57 adult models
and 45 animation states that mix locomotion, stops, turns, and social
interactions. Seam handling is exercised in every split rather than in a
curated subset: \PanoPedSWrapTracks{} tracks wrap around the seam for
\PanoPedSWrapInstanceFrames{} instance-frames.

\paragraph{Platform motion and synchronized capture.}
Camera motion comes from the platforms. The quadruped's gait follows the
distance it travels, its body attitude follows speed and terrain, and the
camera sits on a damped top mast. The drone's attitude responds to
acceleration, banking, gusts, and motor vibration; four sequences add a
stabilizer that levels the camera using only past motion. The visible platform
is masked in every frame instead of being cropped out. At each simulation
step, a frame barrier freezes crowd, animation, platform, and camera while six
cube faces record RGB, instance ID, and radial range, together with root
states and 22 body joints. Each modality is resampled to ERP in its own way
(Appendices~\ref{app:capture}--\ref{app:embodied}). All evaluated trackers use
RGB alone at inference.

\begin{figure}[t]
\centering
\figfile{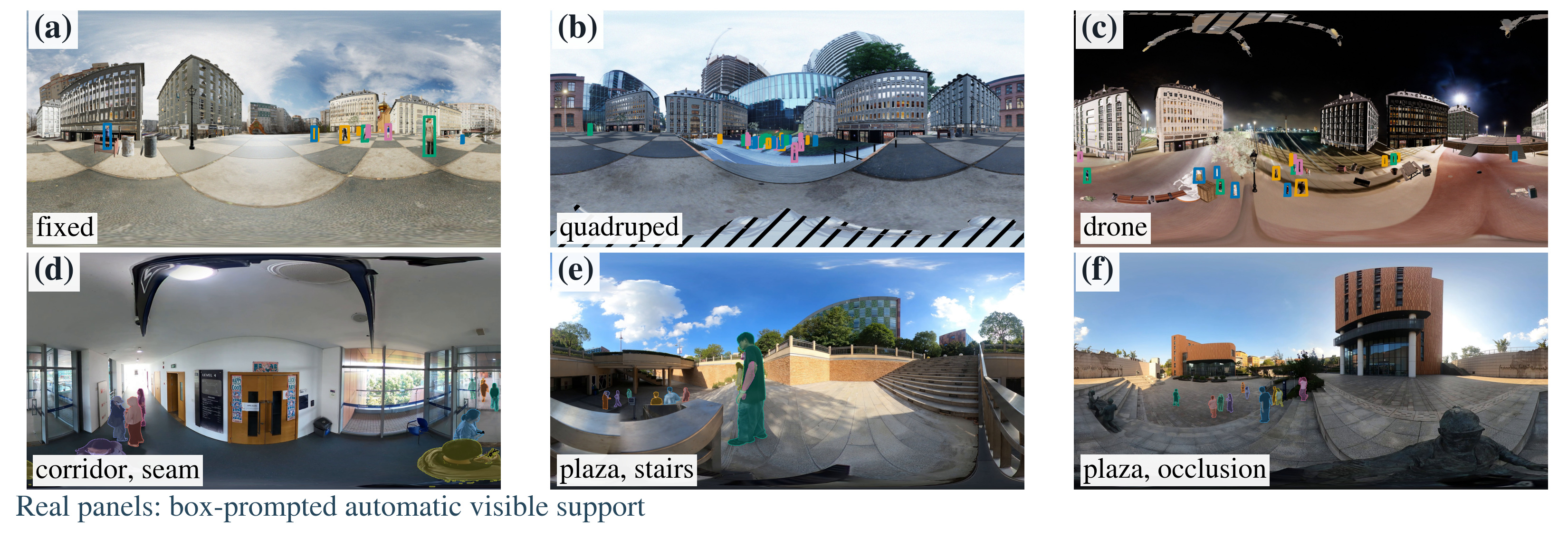}{4.8cm}{}
\caption{\textbf{Both components of \bench{}.}
(a--c) Released \benchS{} frames from a fixed, a quadruped, and a drone rig
with native regions; hatched overlays mark the visible moving platform.
(d--f) \benchR{} frames: a corridor with seam and lower-polar targets,
stairs, and scale variation with occlusion. Colored overlays show
box-prompted automatic visible support from Support-auto v0.1.}
\label{fig:gallery}
\end{figure}

\subsection{\benchR{}: dense real annotation on the full sphere}
\label{sec:panopedr}
\benchR{} was recorded with a GoPro MAX on fixed supports in an indoor
corridor, a corridor corner, a rooftop plaza, and two multilevel plazas with
stairs, and released as gravity-aligned $3840\times1920$ ERP video at
29.97\,fps: \PanoPedRFrames{} frames over \PanoPedRDurationSeconds{}\,s.
Three recordings are densely annotated with \PanoPedRLabeledFrames{} frames,
\PanoPedRTracks{} identities, and \PanoPedRBoxes{} modal boxes, of which
\PanoPedRScoreableBoxes{} are scoreable. Ten consenting participants produce
crossings, occlusion, and re-entry, including \PanoPedRSeamTracks{}
seam-crossing tracks and \PanoPedRPolarBoxes{} lower-polar boxes
(Figure~\ref{fig:gallery}).

The human annotation records visible-body boxes, persistent identities, five
visibility levels rating how clearly a person can be recognized, and
\emph{outside} intervals where a person cannot be reliably located. Adaptive
keyframes, separate identity and geometry passes, and polar-view checks
support dense annotation (Appendix~\ref{app:real_data}).

\paragraph{Real localization targets.}
Besides the Box-fit regions derived from the human boxes, we release
\supportR{}. It obtains a mask for each human box with an automatic segmenter
and converts the mask with the same operator $F_S$ as in S. This layer gives
149,202 valid observations over the three labeled videos and is the target of
our sim-to-real evaluation; images and identities are unchanged. The masks are
automatic and have not been checked by hand. Their construction, coverage, and
the 913 observations without a usable mask are described in
Appendix~\ref{app:support_auto}. Box-fit remains available as a second,
box-based protocol.

\begin{table}[t]
\centering\small
\caption{\textbf{\bench{} at a glance.} Identities are sequence-local
tracks; seam-wrapping counts are tracks whose periodic rectangle crosses the
seam (instance-frames in parentheses).}
\label{tab:dataset_summary}
\setlength{\tabcolsep}{4pt}
\begin{tabularx}{\linewidth}{@{}p{.24\linewidth}XX@{}}
\toprule
 & \benchS{} & \benchR{} \\
\midrule
Source & Unreal Engine 5.8.2, one scene family & GoPro MAX, five viewpoints \\
Platforms & 30 fixed, 18 quadruped, 12 drone & Fixed full-sphere camera \\
Sequences / frames & 60 / 108,000, all labeled & 5 / 28,002; 3 / 16,247 labeled \\
Resolution, rate & 2K (56 clips), 4K (4 clips), 30 fps & $3840\times1920$, 29.97 fps \\
Identity tracks & 1,486 (796 fixed, 690 moving) & 30 (10 per labeled sequence) \\
Seam-wrapping tracks & 799 (40,518) & 10 \\
Native target (L1) & Visible-support centroid and spans & Box-fit: fitted to each box; Support-auto: centroid/spans of automatic masks \\
Supervision & Masks, depth, pose, 3D state, 22 joints & Human boxes/IDs/visibility; automatic support masks \\
Protocol & 44 / 8 / 8 sequence split & Sim-to-real; 3 LOSO folds \\
\bottomrule
\end{tabularx}
\end{table}

\subsection{Training and evaluation protocols}
\label{sec:protocols}
\emph{Synthetic}: 44 training, eight validation, and eight test sequences; the
validation set has four development (V-dev) and four held-back audit (V-audit)
sequences. \emph{Sim-to-real}: train on S and evaluate the frozen model on all
three real videos without fitting any R labels. Our main real-video results
use Support-auto, so that targets are defined by visible pixels in both
domains. \emph{Real-only leave-one-sequence-out (LOSO)}: with only three
labeled sequences in \benchR{}, we do not recommend this setting as a
standalone benchmark. For completeness, Appendix~\ref{app:support_results}
reports the full study: each fold trains on two R sequences and tests on the
third, and the three held-out prediction sets are combined. Both real settings
evaluate 16,247 frames. HOTA averages over 19 native-IoU thresholds, and IDF1
uses a threshold of 0.5 \citep{luiten2021hota}. Metrics are computed from
counts summed over complete sequences, and observations without a usable mask
are handled identically for every readout (Appendix~\ref{app:support_auto}).

\section{\method{}: A Query-Conditioned Localization Head}
\label{sec:method}
\method{} is a small head attached to the output of a trained, frozen tracker.
For each person kept in a frame, the tracker outputs an ERP box $B_i$, a
visual query $q_i\in\mathbb{R}^{256}$ (the detector's feature vector for that
person), a confidence $s_i$, and an identity $k_i$. \method{} reads $B_i$,
$q_i$, and $s_i$ and predicts how to move and resize the person's region on
the sphere; it never changes $k_i$ or $s_i$.

\subsection{From the box to the visible person}
\label{sec:bottleneck}
The usual way to obtain a spherical region is a fixed conversion $G(B_i)$ that
uses only the box. In \benchS{}, however, the target is defined by the
person's visible pixels, the instance mask $M_i$. Its center direction is the
average direction of these pixels, weighted by the area each pixel covers on
the sphere:
\begin{equation}
 u(M_i)=\frac{\sum_{p\in M_i} w_p u_p}
 {\left\|\sum_{p\in M_i} w_p u_p\right\|_2},
 \qquad w_p\propto\cos\varphi_p,
\label{eq:support_centroid}
\end{equation}
where $u_p$ is the unit ray through pixel $p$ and $w_p$ compensates for the
stretching of ERP rows near the poles. The angular width and height of the
visible pixels, measured around this center in a local frame aligned with
gravity, complete the target. The target thus describes the visible part of
the person, not the full body behind occluders.

\begin{figure}[t]
\centering
\figfile{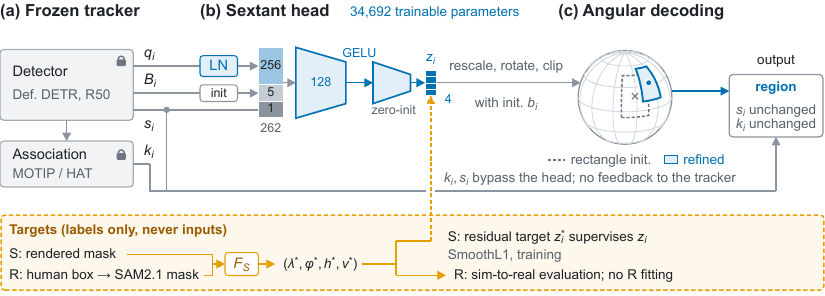}{5.0cm}{}
\caption{\textbf{\method{} is attached after tracking.} (a) The frozen tracker
provides a query, box, score, and identity for each person. (b) The head
normalizes the query, appends the five values of the starting region and the
score, and predicts a four-value correction ($262\to128\to4$ MLP). (c) The
correction moves the center and rescales the angular width and height. Scores
and identities bypass the head, with no feedback to the tracker. Dashed block:
S targets (rendered masks) and R targets (box-prompted automatic masks) share
$F_S$. The head is trained on S and, for sim-to-real transfer, applied to R
without fitting R labels.}
\label{fig:method}
\end{figure}

The box alone cannot recover this target. The geometric readout $G$ treats the
box as uniformly filled and returns its area-weighted center and spans, so two
people in different poses who share one box receive the same region
(Figure~\ref{fig:teaser}b). The query $q_i$, computed from image features
before box regression, still carries information about where the person is
inside the box, and \method{} uses it to learn how the region should differ
from the box (Figure~\ref{fig:method}).

\subsection{Predicting a correction to the box}
\label{sec:head}
\method{} does not predict a region from scratch; it predicts a correction to
a starting region read off the box. The starting region
$b_i=(u_{ix},u_{iy},u_{iz},\log h_i,\log v_i)$ has center direction $u_i$, the
ray through the center of the box (which may wrap around the seam and is
clipped to the image vertically), angular width $h_i$ and height $v_i$ in
radians, and zero roll. It differs from the area-weighted readout $G$. The
head maps the normalized query, $b_i$, and the score to a four-dimensional
correction
\begin{equation}
 z_i=h_\theta\!\left([\LN(q_i);\,b_i;\,s_i]\right)\in\mathbb{R}^4.
\label{eq:head}
\end{equation}
The LayerNorm has 256 channels, and the MLP has widths $262\to128\to4$, one
GELU activation, and a zero-initialized output layer; together they have
\HeadParams{} trainable parameters.

The correction is expressed relative to the starting region. For a target with
center $(\lambda_i^*,\varphi_i^*)$ and spans $(h_i^*,v_i^*)$, the training
target is
\begin{equation}
 z_i^*=\left(\frac{\wrap(\lambda_i^*-\lambda_i)}{h_i},\;
 \frac{\varphi_i^*-\varphi_i}{v_i},\;
 \log\frac{h_i^*}{h_i},\;\log\frac{v_i^*}{v_i}\right),
 \quad \wrap(a)=\operatorname{atan2}(\sin a,\cos a).
\label{eq:target}
\end{equation}
The first two entries are the center shift in longitude and latitude, in units
of the box's own width and height, so the same value means a similar relative
shift for near and far people. The last two are log span ratios, so a
correction can rescale a span but never make it negative. The function $\wrap$
takes the shorter way around the circle, so a person straddling the seam does
not produce a $2\pi$ jump; like any signed angle, it still jumps at $\pm\pi$.

To apply a correction, the decoder first scales the first two entries by $h_i$
and $v_i$ to obtain angles. It then rotates the starting ray up or down about
its local east axis and left or right about the world vertical, and clips the
latitude to $[-\pi/2,\pi/2]$. The spans are rescaled as
\begin{equation}
 \widehat h_i=\min\{2\pi,h_i\exp(z_{i3})\},\qquad
 \widehat v_i=\min\{\pi,v_i\exp(z_{i4})\}.
\label{eq:decode_spans}
\end{equation}
Since the output layer starts at zero, the untrained head returns exactly the
rectangle-center region, not the output of $G$, so every change from it is
learned.

\subsection{Training and use with a frozen tracker}
\label{sec:training}
For training, we match the frozen detector's outputs one to one with the L1
targets of the 44 training sequences once, in advance, and keep the pairs with
native IoU $\ge0.5$. With minibatches $\mathcal M$ that weight all training
sequences equally, we minimize
\begin{equation}
 \mathcal L_{\mathrm{loc}}(\theta)=\frac{1}{|\mathcal M|}
 \sum_{i\in\mathcal M}\operatorname{SmoothL1}(z_i-z_i^*).
\label{eq:loc_loss}
\end{equation}
We use batch size 256, 25,000 AdamW updates, learning rate $3\times10^{-4}$,
weight decay $10^{-4}$, and gradient clipping at 1, and train two heads (seeds
42 and 43) on the same frozen detector. Only matched outputs are used in
training, but every tracker output is refined at inference, where the output
at frame $t$ becomes
\begin{equation}
 \widehat{\mathcal T}_t=\{(k_i,s_i,\operatorname{Decode}(b_i,z_i))\}_i,
 \qquad (k_i,s_i)=(k_i^{\mathrm{base}},s_i^{\mathrm{base}}),
\label{eq:plug_in}
\end{equation}
so each output keeps its identity and score and receives a refined region.
\method{} needs only the per-detection queries: it adds no image encoder,
retrains no identities, and keeps no state over time. Its weights are tied to
the detector and the target definition used in training. Because HAT runs on
the same frozen detector as MOTIP, one S-trained head serves both trackers
without changing its inputs. Section~\ref{sec:real} tests transfer to real
video, and Appendix~\ref{app:support_results} reports training on real data.

Three controls each remove one ingredient. The \emph{ERP-supervised query
head} sees the same query but refines the ERP box, which $G$ then converts; it
tests whether the native target matters. The \emph{geometry-only head} keeps
the native target but sets $q_i=0$, testing whether image evidence is needed.
The \emph{capacity-matched geometry head}, a $6\to3153\to4$ MLP with 34,687
active parameters, tests whether parameter count alone explains the gain,
although the two networks still differ in form.

\section{Experiments}
\label{sec:experiments}

\subsection{Setup}
\label{sec:setup}
Our shared base is MOTIP with Deformable DETR and ResNet-50
\citep{gao2025motip,zhu2021deformabledetr}, initialized from COCO, trained for
20 detection epochs on \benchS{} and then jointly for tracking. Among the
checkpoints after three, four, and five joint epochs (e3, e4, e5), we select
e3 by V-dev HOTA with the geometric readout. ByteTrack
\citep{zhang2022bytetrack} and OC-SORT \citep{cao2023ocsort} use the same
detections. HAT \citep{gao2025hat} adds MASA-R50 features \citep{li2024masa},
and its history-off control keeps these features but disables the history
transform. D-FINE-L with OC-SORT \citep{peng2025dfine} and OmniTrack
\citep{luo2025omnitrack} use their own detectors. All methods use the released
geometry and ignore rules. Appendix~\ref{app:budgets} gives the training
budgets and the checkpoint selection; the comparison does not assume identical
pretraining or training budgets.

\subsection{Results on the \benchS{} test set}
\label{sec:main_results}
Table~\ref{tab:final_test} reports the complete test comparison. MOTIP with
the geometric readout is the strongest baseline in this comparison, ahead
of HAT by 2.56 HOTA. \method{}
raises it to 49.49 and 49.47 HOTA and 56.06 and 56.05 IDF1 for the two
seeds, improving on the strongest baseline by 2.19 and 2.17 HOTA. Both heads
improve HOTA and IDF1 on every one of the eight test sequences
(Figure~\ref{fig:analysis}c), and 2,000 paired whole-sequence bootstrap
resamples give 2.5--97.5 percentile ranges of $[1.58,2.79]$ and $[1.54,2.75]$
for the HOTA gain. The improvement does not come from identity: output
identities and scores are byte-identical to those of the frozen tracker.

\begin{table}[t]
\centering
\caption{\textbf{\benchS{} test set: eight sequences, 14,400 frames.}
$\Delta$HOTA is relative to MOTIP with the geometric readout. Systems above
the first rule train their own detector; systems below share the frozen
MOTIP e3 detector. Every system except ours forwards its rectangle through
the geometric readout. \textbf{Bold}: best; \underline{underlined}: second.}
\label{tab:final_test}
\setlength{\tabcolsep}{5pt}
\footnotesize
\begin{tabular}{@{}lllrrr@{}}
\toprule
Method & Venue & Detector & HOTA & IDF1 & $\Delta$HOTA \\
\midrule
OmniTrack & CVPR'25 & own & 21.18 & 20.20 & \loss{$-26.12$} \\
D-FINE + OC-SORT (42) & ICLR'25 & own & 27.81 & 23.28 & \loss{$-19.49$} \\
D-FINE + OC-SORT (43) & ICLR'25 & own & 29.75 & 24.80 & \loss{$-17.55$} \\
\midrule
ByteTrack & ECCV'22 & shared & 35.99 & 35.00 & \loss{$-11.31$} \\
OC-SORT & CVPR'23 & shared & 34.45 & 32.99 & \loss{$-12.85$} \\
HAT, history off & control & shared & 42.48 & 47.11 & \loss{$-4.82$} \\
HAT + MASA & ECCV'26 & shared & 44.74 & 51.38 & \loss{$-2.56$} \\
MOTIP, geometric readout \citep{gao2025motip} & CVPR'25 & shared & 47.30 & 55.70 & -- \\
\midrule
\ours\textbf{MOTIP + \method{} (ours), seed 42} & \ours & \ours shared & \ours\textbf{49.49} & \ours\textbf{56.06} & \ours\gain{$\mathbf{+2.19}$} \\
\ours\textbf{MOTIP + \method{} (ours), seed 43} & \ours & \ours shared & \ours\underline{49.47} & \ours\underline{56.05} & \ours\gain{$\underline{+2.17}$} \\
\bottomrule
\end{tabular}
\end{table}

The other rows are also informative. Transforming the same MASA features with
track history gives HAT 2.3 HOTA and 4.3 IDF1, but HAT stays below MOTIP,
whose identity decoder uses the history directly. Detector accuracy predicts
tracking poorly. D-FINE reaches a higher AP75 than the MOTIP detector (59.29
and 60.78 versus 53.76 on V-dev) but lower tracking scores, because precise
boxes do not settle which detections are kept or how they are linked over
time. OmniTrack is less effective in our full-sphere reproduction; it was
designed for annular images, and its reported training budget differs from
ours.

\subsection{Ablation of localization gains}
\label{sec:analysis}
\paragraph{Visual evidence and native supervision.}
Table~\ref{tab:controls} holds the detector and identity stream fixed and
varies only the readout on all three splits. The ERP-supervised query head
sees the same features as \method{} yet moves no score by more than 0.07
HOTA, so refining an ERP rectangle under a matched budget does not reproduce
the native-target gain. The geometry-only head and its capacity-matched
counterpart, which see the native target but not the image, recover only
0.20--0.34 HOTA on test, so generic box calibration does not explain the
improvement either. Only native supervision with the visual query produces
the full 1.65--2.19 HOTA across the three splits, with at most 0.06 HOTA
between seeds.

\begin{table}[t]
\centering\footnotesize
\caption{\textbf{Readout ablation on PanoPed-S with fixed detector and identity outputs.}
HOTA for seeds 42\,/\,43. The query and target columns identify the input
and supervision changes. Full HOTA/IDF1 results are in
Table~\ref{tab:controls_full}.}
\label{tab:controls}
\setlength{\tabcolsep}{3pt}
\begin{tabular}{@{}lccrrr@{}}
\toprule
Readout & Query & Target & V-dev & V-audit & Test \\
\midrule
Area centroid & \xmark & -- & 49.06 & 41.41 & 47.30 \\
ERP query & \cmark & ERP & 49.09/49.12 & 41.39/41.41 & 47.24/47.26 \\
Geometry-only & \xmark & native & 49.27/49.21 & 41.77/41.81 & 47.61/47.50 \\
Matched-capacity geometry & \xmark & native & 49.09/49.32 & 41.63/41.83 & 47.52/47.64 \\
\ours\textbf{\method{}} & \ours\cmark & \ours native & \ours\textbf{51.03/51.04} & \ours\textbf{43.12/43.06} & \ours\textbf{49.49/49.47} \\
\bottomrule
\end{tabular}
\end{table}

\begin{figure}[t]
\centering
\figfile{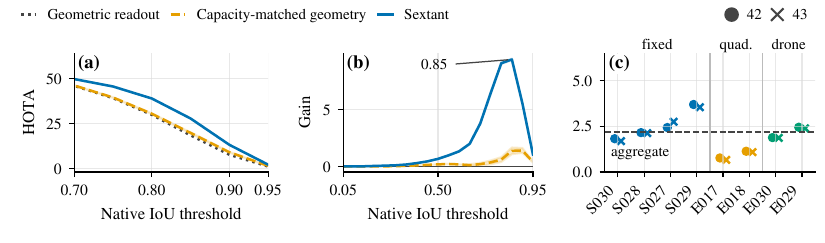}{4.1cm}{}
\caption{\textbf{Localization analysis on the test set.} (a) HOTA at native-IoU
thresholds 0.70--0.95 for the geometric readout, capacity-matched geometry
heads, and \method{}. (b) Gain over the geometric readout at all 19 thresholds, which widens under
stricter matching. (c) HOTA gain of \method{} on each test sequence for both
seeds (circle: 42; cross: 43), grouped as fixed (blue), quadruped (orange),
and drone (green). S/E abbreviate SS/SE. The dashed line marks the mean
aggregate gain of the two seeds; every sequence improves.}
\label{fig:analysis}
\end{figure}

\paragraph{Geometric effects.}
On 117,875 fixed V-dev matches, mean angular error between predicted and
annotated center rays falls from $0.524^\circ$ for rectangle-center conversion
to $0.412^\circ$ and $0.434^\circ$ for \method{} seeds 42 and 43.
This diagnostic uses the direct rectangle reference, not the area-centroid
baseline in the tracking tables. The reported mean absolute horizontal and
vertical log-span errors decrease from $0.148$ and $0.042$ to $0.087$ and
$0.036$, including a 41\% reduction in the horizontal term.
Figure~\ref{fig:analysis}a,b shows larger tracking gains at stricter matching
thresholds, peaking at 0.85. Fixed-rule qualitative examples appear in
Appendix~\ref{app:qualitative}.

\paragraph{Identity metrics, endpoints, and cost.}
Better overlap changes which predictions the evaluator matches, so IDF1 rises
by 0.36 with an unchanged identity stream while the ID-switch count moves
from 7,514 to 7,608 and 7,627. Heads trained for the e4 endpoint improve
V-dev HOTA from 45.74 to 47.61 and 47.46, so the effect is not tied to one
checkpoint, and measured complete-pipeline throughput remains comparable at 11.1--11.7 fps
on RTX 5090 D and 6.0--6.2 fps on RTX 4090 D (Appendix~\ref{app:runtime}).

\subsection{Sim-to-real tracking without fine-tuning}
\label{sec:real}
On all 16,247 real frames (Table~\ref{tab:support_source}), the frozen
S-trained heads improve MOTIP by 0.96--1.13 HOTA and HAT by 0.97--1.14.
Both seeds also improve IDF1 and outperform their geometry controls.

\begin{table}[t]
\centering\small
\caption{\textbf{Sim-to-real tracking: all 16,247 real frames.}
Support-auto targets; no R fine-tuning or added localization encoder.
$\Delta$: gain over the same tracker's $U$. Geometry and \method{} add
34,687 and 34,692 parameters. Bold/underline: best/second within each
tracker. Extra-encoder comparisons: Appendix~\ref{app:support_results}.}
\label{tab:support_source}
\setlength{\tabcolsep}{4pt}
\begin{tabular}{@{}llrrrr@{}}
\toprule
Tracker & Readout & HOTA & IDF1 & $\Delta$HOTA & $\Delta$IDF1 \\
\midrule
\multirow{5}{*}{\shortstack[l]{MOTIP\\CVPR'25}} & Uniform support $U$ & 31.604 & 31.960 & -- & -- \\
 & Geometry 42 & 32.187 & 32.082 & +0.583 & +0.122 \\
 & Geometry 43 & 32.321 & 32.177 & +0.717 & +0.217 \\
 & \ours\textbf{\method{} 42} & \ours\textbf{32.739} & \ours\underline{32.233} & \ours\gain{\textbf{+1.135}} & \ours\gain{\underline{+0.273}} \\
 & \ours\textbf{\method{} 43} & \ours\underline{32.561} & \ours\textbf{32.235} & \ours\gain{\underline{+0.958}} & \ours\gain{\textbf{+0.275}} \\
\midrule
\multirow{5}{*}{\shortstack[l]{HAT\\ECCV'26}} & Uniform support $U$ & 31.711 & 31.948 & -- & -- \\
 & Geometry 42 & 32.302 & 32.048 & +0.591 & +0.100 \\
 & Geometry 43 & 32.453 & 32.215 & +0.742 & +0.267 \\
 & \ours\textbf{\method{} 42} & \ours\textbf{32.850} & \ours\underline{32.276} & \ours\gain{\textbf{+1.139}} & \ours\gain{\underline{+0.328}} \\
 & \ours\textbf{\method{} 43} & \ours\underline{32.679} & \ours\textbf{32.279} & \ours\gain{\underline{+0.968}} & \ours\gain{\textbf{+0.331}} \\
\bottomrule
\end{tabular}
\end{table}

HAT+\method{} leads the compared sim-to-real systems without an additional
localization encoder, including HAT-off (32.59 HOTA). Both seeds improve
every real sequence with either tracker. Per-sequence results,
automatic-label sensitivity, and comparisons that add a segmentation encoder
are collected in Appendix~\ref{app:support_results}.

\section{Conclusion}
\label{sec:conclusion}
To our knowledge, \bench{} is the first sim-to-real benchmark for full-sphere
pedestrian tracking, pairing 108,000 synthetic and 16,247 densely annotated
real frames under one visible-support target rule. These targets expose
localization lost when trackers reduce visual features to boxes.
\method{}, a roughly 0.035M-parameter query readout, raises MOTIP from 47.30
to 49.49 HOTA on the \benchS{} test set without tracker retraining.
Synthetic-only training also improves MOTIP and HAT by 0.96--1.14 HOTA on
real video, with gains on every sequence. The synthetic pipeline can supply
spherical labels for future sphere-native detectors.
\label{maintext:end}

\clearpage
\subsection*{Ethics statement}
The real recordings were collected with institutional ethics permission and
written informed consent from all ten participants. Capture locations are
described by scene type only. \benchR{} is distributed under the \benchR{} Research-Only Data
License v0.9.0, which permits noncommercial research and education and
prohibits identification, re-identification, and surveillance uses.
\benchS{} contains flattened renders and numerical annotations rather than
source meshes, textures, or engine packages, and is distributed under
the PanoPed-S Research-Only Data License v1.0.0, with the upstream asset terms recorded in
Appendix~\ref{app:release}. The real image panels show consented original frames; synthetic panels are unmodified.

\subsection*{Reproducibility statement}
Section~\ref{sec:dataset} and Appendices~\ref{app:dataset}--\ref{app:real_data}
specify the annotation levels, the target operators for both components, the
capture pipeline, the crowd and platform motion models, and quality-control procedures. Section~\ref{sec:method} gives the complete architecture, target
parametrization, and training recipe of \method{}, and
Appendix~\ref{app:experiments} records the development chronology, the
detector selection rule, baseline budgets and fixed tracker settings, the
bootstrap procedure, per-sequence results, and runtime measurement scope. The
release includes annotation transformation and verification scripts,
sequence configurations, frozen model and prediction manifests with hashes,
and the evaluation code. The project page is \url{https://zhuqinfeng1999.github.io/PanoPed/}, and code, dataset download links, and release-specific licenses are available at \url{https://github.com/zhuqinfeng1999/PanoPed}.

\subsection*{AI use statement}
In this work, we used GPT and Claude to polish the language and
improve the readability of the manuscript.

OpenAI Codex assisted with research execution through code
implementation and debugging, including revisions to Python
figure-generation scripts to improve figure presentation.

GPT also assisted with drafting implementation descriptions for
the supplementary material based on the actual source code.

For \benchS{}, Codex assisted with implementing the Unreal Engine
simulation and capture pipeline. The synthetic RGB frames were
rendered in Unreal Engine, not directly synthesized by generative
image or video models. Asset provenance is documented in
Appendix~\ref{app:release}.
For \benchR{}, annotation used ByteTrack-assisted pre-annotation and
manual review in CVAT; the retained initialization records and their
sequence coverage are described in Appendix~\ref{app:real_annotation_process}.
SAM2.1~\citep{ravi2025sam2}, prompted by human-annotated boxes,
produced masks for \supportR{} v0.2, while
SAM-HQ~\citep{ke2023samhq} supplied alternative masks for sensitivity
comparisons. These masks are automatic estimates, not independently
human-verified segmentation ground truth.

We have reviewed all AI-assisted work and take responsibility for
the final content, including the text, code, dataset annotations,
and scientific claims.

\begingroup
\raggedright
\bibliography{references}
\bibliographystyle{iclr2027_conference}
\endgroup
\clearpage
\appendix
\raggedbottom
\section*{Supplementary material}
Appendix~\ref{app:dataset} describes the annotation formats and the data
splits. Appendices~\ref{app:capture} and~\ref{app:embodied} explain how the
synthetic data are rendered and how the crowds and cameras move.
Appendix~\ref{app:real_data} covers how the real videos were recorded and
annotated, and Appendix~\ref{app:release} summarizes what is released and
under which terms. Appendix~\ref{app:experiments} contains the additional
experiments: synthetic controls, training on real data, sim-to-real results,
sensitivity to the automatic masks, and runtime.

\section{Annotation Formats and Conversion Rules}
\label{app:dataset}
\paragraph{Sequences and identities.}
Each synthetic sequence is one continuous 60-second clip, with frames numbered
from one without gaps. No sequence is split in time between training and
evaluation. Identities start at one and are never reused within a sequence. A
person keeps the same identity through a complete occlusion as long as they
are still present in the simulation behind the occluder. People appear and
disappear only at fixed entry and exit points placed in the scene, or when
they move beyond a set distance from the camera. The \PanoPedSIdentityTracks{}
released synthetic tracks count identities within each sequence; they are not
\PanoPedSIdentityTracks{} different-looking people.

\paragraph{Visible pixels and visibility.}
The synthetic L0 masks and L2 boxes describe only what is visible. When a
person is completely hidden but their position is known from the simulation,
we keep a simple capsule-shaped placeholder at that position, set its
visibility to 0.0, and mark it as not scored. Visibility is computed from how
much of the person's own mask appears around their rendered body joints, and
it is rounded to the same five levels used for the real annotations. We keep
the unrounded ratio for analysis, but it never replaces the human judgment
used in \benchR{}, which rates how clearly a person can be recognized.

\paragraph{Boxes that wrap around the seam.}
Because the ERP image wraps around horizontally, the horizontal extent of a
synthetic L2 box is the shortest arc that contains all of the person's
columns. We find it by treating the columns as a circle and removing the
longest run of empty columns; ties are broken by a fixed rule. If this arc
crosses the seam, it is split into two L3 rows with the same identity. Each
piece covers exactly its part of the arc, including empty columns next to the
seam, so the two pieces together give back the periodic box. We also provide
the conventional export with a single scored piece per person; it carries less
spatial information and is labeled separately in the interface.

\paragraph{What S and R share.}
\benchS{} and \benchR{} use the same coordinate conventions, the same format
for identities within a sequence, and the same vocabulary for visibility and
\emph{outside} states. Their L0 and L1 are built differently. Synthetic masks
give the center of the visible area directly, whereas the original real
annotations are boxes, so the Box-fit layer fits a region to each box boundary.
Support-auto adds estimated masks for the real data and converts them with the
same operator as S. Every target describes what is visible in the image, as a
region on the sphere; none is a metric 3D body center or an estimate of the
hidden full body. Using the same operator does not make the automatic real
masks exact.

\begin{table}[H]
\centering\small
\caption{\textbf{Composition of the synthetic splits and reuse of assets.}
Each count is the number of distinct items in a split; the last column
counts items shared by TRAIN/VAL, TRAIN/TEST, and VAL/TEST.}
\begin{tabular}{@{}lrrrr@{}}
\toprule
Item & TRAIN & VAL & TEST & Shared by pairs \\
\midrule
Sequences & 44 & 8 & 8 & 0 / 0 / 0 \\
Scene families & 1 & 1 & 1 & 1 / 1 / 1 \\
Scene layouts & 35 & 7 & 7 & 3 / 3 / 1 \\
Adult avatar models & 57 & 57 & 57 & 57 / 57 / 57 \\
HDRI assets & 13 & 7 & 5 & 7 / 5 / 3 \\
Moving-camera route configurations & 21 & 4 & 4 & 0 / 0 / 0 \\
\bottomrule
\end{tabular}
\end{table}
The splits separate sequences, but they share the scene family, the texture
set, the avatar pool, and the controllers of both moving platforms. The splits
use 5, 4, and 4 fixed camera sites, of which 4, 4, and 3 are shared by
TRAIN/VAL, TRAIN/TEST, and VAL/TEST; no exact combination of camera position
and height appears in two splits. All 79,200 TRAIN frames and 14,400 VAL
frames are 2K. TEST contains 7,200 4K frames from fixed cameras and 7,200 2K
frames from moving cameras. The same ten participants appear in all three
labeled real sequences, so real LOSO holds out sequences, not people. For the
ignore polygon of a fixed mount, pixel $(i,j)$ is inside if its integer index
lies inside the polygon or on its boundary. The half-pixel offset of
Equation~\ref{eq:erp_sphere} is used only when pixels are converted to rays.
The training data loader checks this convention against two reference bitmaps.

\section{Synthetic Capture and Synchronized Labels}
\label{app:capture}
\begin{figure}[H]
\centering
\figfile{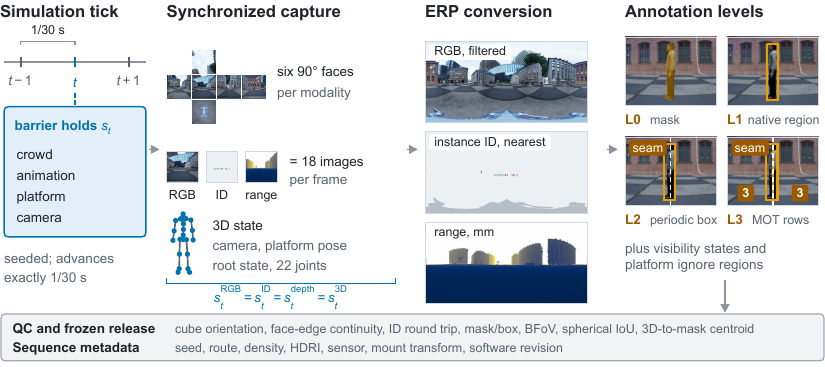}{6.2cm}{}
\caption{\textbf{Synchronized generation of images and labels.} The seeded
simulation advances by exactly $1/30$\,s per frame. A frame barrier holds one
world state $s_t$ while the RGB, instance-ID, range, and 3D pose data are
recorded, so $s_t^{\mathrm{RGB}}=s_t^{\mathrm{ID}}=
s_t^{\mathrm{depth}}=s_t^{\mathrm{3D}}$. The images are then converted to ERP
offline, with a resampling method suited to each modality, and all
annotations, checks, and release files are built from these synchronized
outputs. Every frame is linked to its sequence settings: seed, route, crowd
density, HDRI, sensor, mount transform, and software version. Cube faces are shown by reprojecting the released ERP frame.
Annotation crops are circularly shifted for display so that the seam is interior.}
\label{fig:pipeline}
\end{figure}

We render \benchS{} in \PanoPedSEngine{} and control it with Python code built
on SPEAR's begin-frame/end-frame loop \citep{roberts2022spear}. For each
frame, a barrier freezes the crowd, animations, platform, and camera. Six
$90^\circ$ cube faces are then rendered for RGB, integer instance ID, and
radial range, 18 face images in total. The 3D states of the camera, the
platform, and all pedestrians are logged at the same simulation instant
(Figure~\ref{fig:pipeline}). Each pedestrian record stores root position,
orientation quaternion, velocity, height, behavior and animation state, and 22
named body joints in world coordinates, taken after inverse kinematics (IK)
has adjusted the pose. Camera and platform poses use the same coordinate
convention.

The controller advances the simulation by exactly $1/30$\,s between frames.
Animation, exposure, level of detail, and platform state are not updated while
the modalities of one frame are rendered. RGB is rendered with supersampling
and then filtered, whereas instance IDs are always resampled with nearest
neighbor so that ID values are never blended. A fixed table specifies how the
cube faces map to the ERP image. We check cube orientation, continuity across
face edges, and exact round trips of integer IDs. We also check that ERP masks
and boxes agree, that the conversion to bounding fields of view (BFoV) and
spherical IoU are correct, and that each person's 3D position agrees with the
center of their mask. Depth is the distance from the camera along each pixel's
ray, in millimeters; the largest integer value is reserved for sky,
out-of-range, and invalid pixels. For moving platforms, the ignore region of
each frame is the platform's rendered silhouette; a fixed mount uses one
region for the whole sequence. No tracking experiment in this paper uses
ground-truth depth or pose at inference.

\paragraph{Quality control and human review.}
\label{app:qc}
Before release, automatic checks confirm that every frame and modality was
captured, that identities are integers, and that the annotation levels agree
geometrically. They also check that tracks are continuous in time, body poses
are valid, platforms keep clear of the scene and the crowd, the crowd matches
its specification, and every asset has a recorded source. Human reviewers then
inspect overlays of masks, periodic boxes, native region boundaries,
identities, and visibility; for moving cameras they also check the ignore
region and the camera attitude. We fixed the controller settings and the
acceptance criteria before accepting the random seeds of the released
sequences. Automatic checks and visual review catch different kinds of errors,
so both are required.

\section{Crowds and Camera Motion}
\label{app:embodied}
\subsection{Population and path design}
The synthetic scene is a modern city environment that we rebuilt, with a lower
plaza, an upper platform, two flights of stairs at human scale, long and short
lines of sight, and street furniture. Thirteen map and HDRI variants change
the lighting, materials, and viewpoints within this scene family. Five fixed
sites place the camera 0.8--3.0\,m above the ground. We target 5--50
pedestrians for fixed cameras and 16--32 for moving cameras. The release
contains \PanoPedSStaticIdentityTracks{} tracks from fixed cameras and
\PanoPedSEgoIdentityTracks{} from moving cameras, counted within each
sequence.

\begin{figure}[H]
\centering
\figfile{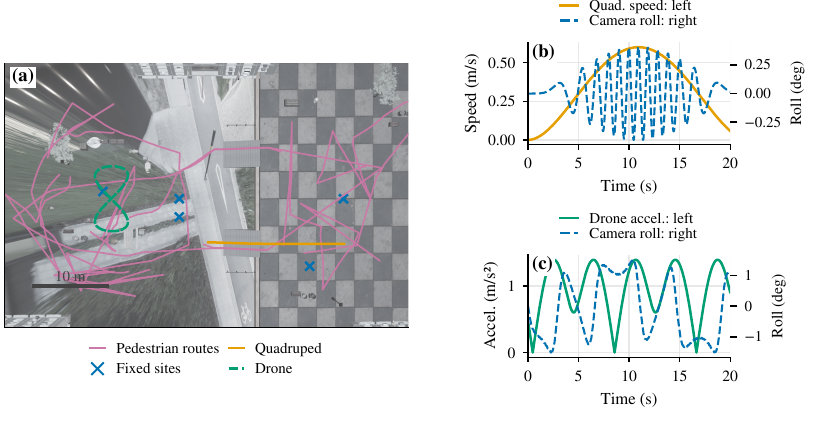}{7.2cm}{}
\caption{\textbf{Where people walk and how the cameras move.} (a) Released
trajectories drawn on an overhead render of the scene. For display only, we
divide the scene into a fixed $4\times2$ grid and show the longest suitable
pedestrian path in each cell; every cell contains one, giving four paths in
each half. Platform paths come from the released logs of SE-012 and SE-026,
and crosses mark the five fixed sites. (b) Quadruped speed and the resulting
camera roll in SE-012. (c) Commanded drone acceleration and the rendered roll
in SE-026 with stabilization. The curves in (b, c) are computed from the fixed
controller settings, not read from camera logs. The selection for display does
not affect which sequences are in the benchmark.}
\label{fig:motion_design}
\end{figure}

\paragraph{Crowd behavior.}
Pedestrians move with the engine's navigation system and Detour crowd
avoidance, using a 0.3\,m radius. Each identity draws its walking speed from a
log-normal distribution with $\mu=\log1.15$ and $\sigma=0.42$, clipped to
$[0.5,2.2]$\,m/s. Routes connect entry and exit points on the lower and upper
levels through paths on the same level, stairs used in both directions, stops,
and meeting points where people gather. The adult avatars and the motion
library come from RocketBox \citep{gonzalezfranco2020rocketbox}. The walking
animation is driven by the distance traveled, with blended turns and standing
states for looking around, using a phone, and talking or listening. Released
sequences use 14--20 standing or social animations, and no single animation
accounts for more than 14.26\% of standing samples.
Figure~\ref{fig:motion_design} shows the released trajectories and the camera
motion that follows from the platform state. Inverse kinematics places each
foot on the stairs, which have 30\,cm treads and 15\,cm risers. Automatic
checks reject sudden jumps in position, changes in body scale, long periods of
walking in place, feet sinking into the ground, and planted feet that slide
too far.

\begin{table}[H]
\caption{\textbf{Routes of the moving cameras.} ``Raw'' keeps the physical
camera attitude; ``stabilized'' only adds the gravity stabilizer described
below, which uses past motion alone.}
\label{tab:sego_routes}
\centering\small
\begin{tabular}{llrr}
\toprule
Platform & Route & Raw & Stabilized \\
\midrule
Quadruped & Plaza walk / trot / stop / turn & 8 & 0 \\
Quadruped & Ramps and stairs across levels & 6 & 0 \\
Quadruped & Curbs, obstacles, speed changes & 4 & 0 \\
Drone & Low-altitude translation and orbit & 4 & 0 \\
Drone & Mid-altitude figure-eight & 2 & 2 \\
Drone & Vertical take-off, hover, landing & 1 & 1 \\
Drone & Fast banked spiral turn & 1 & 1 \\
\midrule
Total & & 26 & 4 \\
\bottomrule
\end{tabular}
\end{table}

\begin{figure}[H]
\centering
\includegraphics[width=\linewidth]{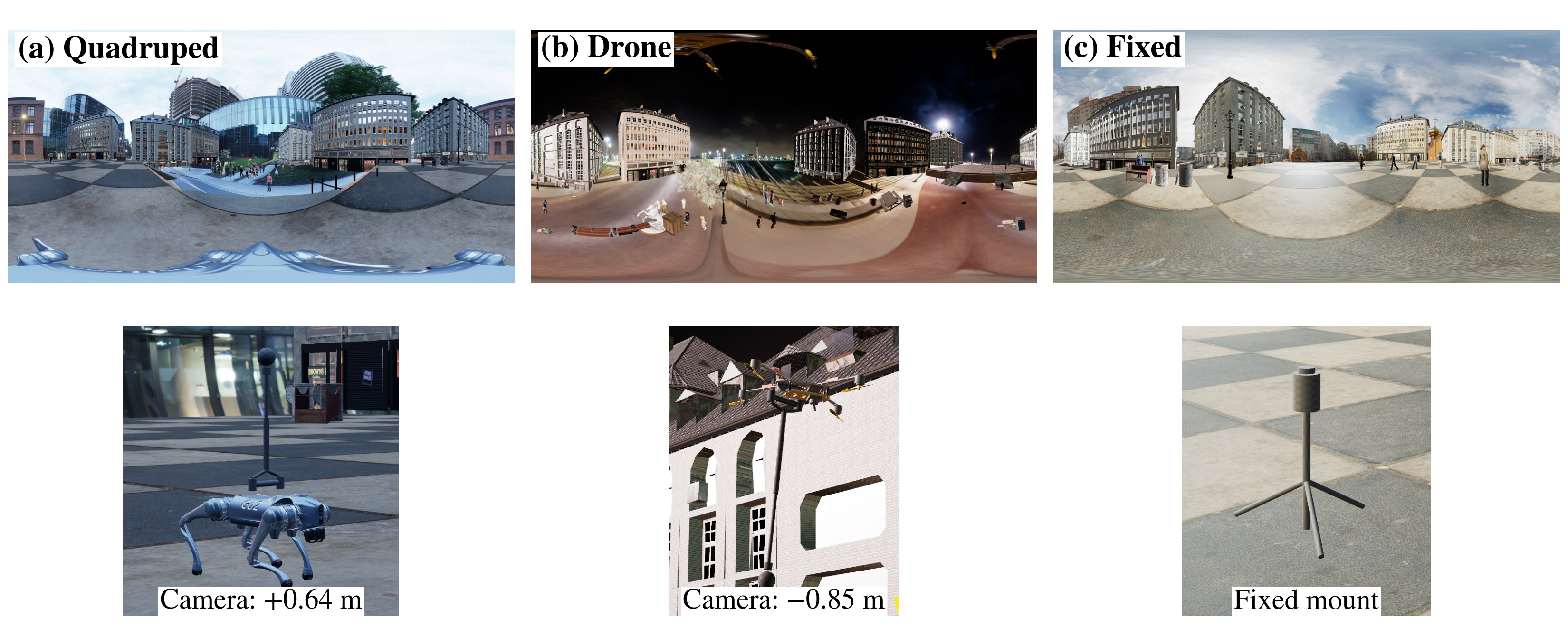}
\caption{\textbf{The three camera platforms in \benchS{}.} (a) Quadruped, (b)
drone, and (c) fixed rig. Each upper panel is a released ERP frame. Each lower
panel shows the same scene and the same platform pose from a third-person
view, with the crowd hidden, to verify the setup. The quadruped's top mast
holds the camera 0.64\,m above its mount and the drone's belly mast 0.85\,m
below it; platform pixels in the released images are covered by the
corresponding ignore regions.}
\label{fig:capture_rigs}
\end{figure}

\paragraph{Quadruped camera.}
The rendered Unitree Go2 model comes from MuJoCo Menagerie (BSD-3-Clause);
only rendered RGB images and numerical annotations are released. The camera
moves because the body responds to speed and because the mast is damped. The
gait phase advances with the distance traveled, so the legs stop when the
robot stops instead of cycling in place. One cycle of the diagonal leg pairs
corresponds to 0.56\,m of travel of the same foot, and we check the rendered
silhouettes for smooth motion where the phase wraps around.
Table~\ref{tab:sego_routes} lists the routes of both moving platforms, and
Figure~\ref{fig:capture_rigs} shows each mount together with a third-person
view rendered for checking. Routes are closed loops through waypoints,
smoothed to minimize jerk, with target speeds of 0.55--1.50\,m/s, turns
lasting 2.2--3.5\,s, and optional stops of 0--0.8\,s. Speed, gait phase, foot
contact, and the slope of the ground drive the body motion, which has a
vertical component of 1.2\,mm and pitch, roll, and yaw components of
$0.6^\circ$, $0.95^\circ$, and $0.3^\circ$. The mast responds at 4\,Hz with
damping ratio 0.75 and holds the camera 0.64\,m above the platform frame. All
18 sequences pass checks for gait continuity, collisions, ground contact,
camera placement, annotations, and whether the view is usable.

\paragraph{Drone camera.}
The drone state consists of position, attitude, velocity, angular velocity,
and rotor phase. The attitude responds to bounded acceleration, to the yaw
rate, to gusts, and to motor vibration, and the camera hangs 0.85\,m below the
airframe. The controller limits the bank angle to $22^\circ$. Four gust
components act with scales of $0.9^\circ$ in attitude and 0.015\,m in
position, and the motor phase advances at 96\,Hz, adding $0.18^\circ$ of motor
vibration and $0.12^\circ$ of yaw vibration. Speed is limited to 4\,m/s,
acceleration to 4\,m/s$^2$, jerk to 12\,m/s$^3$, angular speed to
$120^\circ$/s, and angular acceleration to $500^\circ$/s$^2$. Four sequences
use a gravity stabilizer that relies only on past motion ($\tau=0.35$\,s, at
most $25^\circ$ of correction); eight keep the raw physical orientation. The
airframe stays visible in the panorama and is covered by an ignore mask that
is updated in every frame.

\begin{table}[H]
\caption{\textbf{Continuity check of the drone silhouette in all 12 released
sequences.} Each row measures, on the sphere, how much the drone's rendered
silhouette changes between frames. Frames are compared at the same rotor
phase, so the normal motion of the blades is not counted as a jump. All
sequences also pass checks for collisions, flying over people, labels, and
usable views.}
\label{tab:drone_review}
\centering\small
\begin{tabular}{lrrr}
\toprule
Quantity & Limit & Release max & Worst sequence \\
\midrule
Phase-class XOR excess & 2.0\% & 1.920\% & SE-028 \\
Rotor-wrap XOR excess & 2.0\% & 1.596\% & SE-030 \\
Phase-aligned solid angle & 2.5\% & 1.967\% & SE-025 \\
Spherical second moment & 2.0\% & 0.818\% & SE-026 \\
Spherical centroid & $0.5^\circ$ & $0.340^\circ$ & SE-026 \\
\bottomrule
\end{tabular}
\end{table}

The check in Table~\ref{tab:drone_review} covers all $12\times1{,}800$ drone
frames. For the raw and stabilized versions of the same flight, the platform
position, platform attitude, and raw camera motion are byte-identical;
stabilization changes only the recorded camera attitude. We fixed the
controller settings and acceptance criteria before accepting or rejecting the
random seeds of the released sequences.

\section{Real-Data Acquisition and Annotation}
\label{app:real_data}
\subsection{Acquisition and preprocessing}
\label{app:real_capture}
\paragraph{Capture design.}
We recorded ten consenting participants with a GoPro MAX on fixed supports in
a corridor, a corridor corner, a rooftop plaza, and two sunken plazas with
stairs. Participants entered, paused, crossed, and re-entered in organized
groups; their timing and interactions were not scripted frame by frame.
Identities are defined within each sequence, and the dataset does not link the
same person across sequences. The cameras were fixed, but the scenes were not
measured metrically, and one view can include an upper landing, stairs, and a
lower plaza. The annotations are therefore angular and image-based and do not
assume a flat ground.

\begin{table}[H]
\centering\small
\caption{\textbf{Real recordings.} R-01, R-02, and R-03 are labeled; U01 and
U02 contain images only. Durations assume 29.97\,fps. The lower table gives
the horizontal circular shift applied to each video in three units: pixels of
the source video (width 5,376), pixels of the released images (width 3,840),
and degrees. The shift only moves the seam and says nothing about the camera's
orientation. The last two columns give the 95th percentile of the measured
image displacement and its horizontal and vertical peak-to-peak range within a
ten-second interval, both in source pixels.}
\label{tab:real_overview}
\setlength{\tabcolsep}{5pt}
\begin{tabular}{@{}llrr@{}}
\toprule
Sequence & Scene & Frames & Duration (s) \\
\midrule
R-01 & Corridor and passage & 5,486 & 183.05 \\
U01 & Corridor corner & 6,293 & 209.98 \\
U02 & Rooftop plaza & 5,462 & 182.25 \\
R-02 & Plaza and stairs & 6,416 & 214.08 \\
R-03 & Plaza with partial stairs & 4,345 & 144.98 \\
\midrule
Total & Five fixed views & 28,002 & 934.33 \\
\bottomrule
\end{tabular}
\par\medskip
\begin{tabular}{@{}lrrrrr@{}}
\toprule
Sequence & \shortstack{Source shift\\(px)} & \shortstack{Released shift\\(px)} & \shortstack{Shift\\(deg)} & \shortstack{Motion\\95th pct. (px)} & \shortstack{10-s range\\$x/y$ (px)} \\
\midrule
R-01 & 0 & 0 & 0.00000 & 2.57 & 0.278 / 0.270 \\
U01 & 924 & 660 & 61.87500 & 7.37 & 0.159 / 0.204 \\
U02 & 1,750 & 1,250 & 117.18750 & 5.56 & 1.001 / 0.948 \\
R-02 & 3,248 & 2,320 & 217.50000 & 5.10 & 0.185 / 0.195 \\
R-03 & 4,053 & 2,895 & 271.40625 & 2.45 & 0.177 / 0.237 \\
\bottomrule
\end{tabular}
\end{table}

\begin{figure}[H]
\centering
\includegraphics[width=\linewidth]{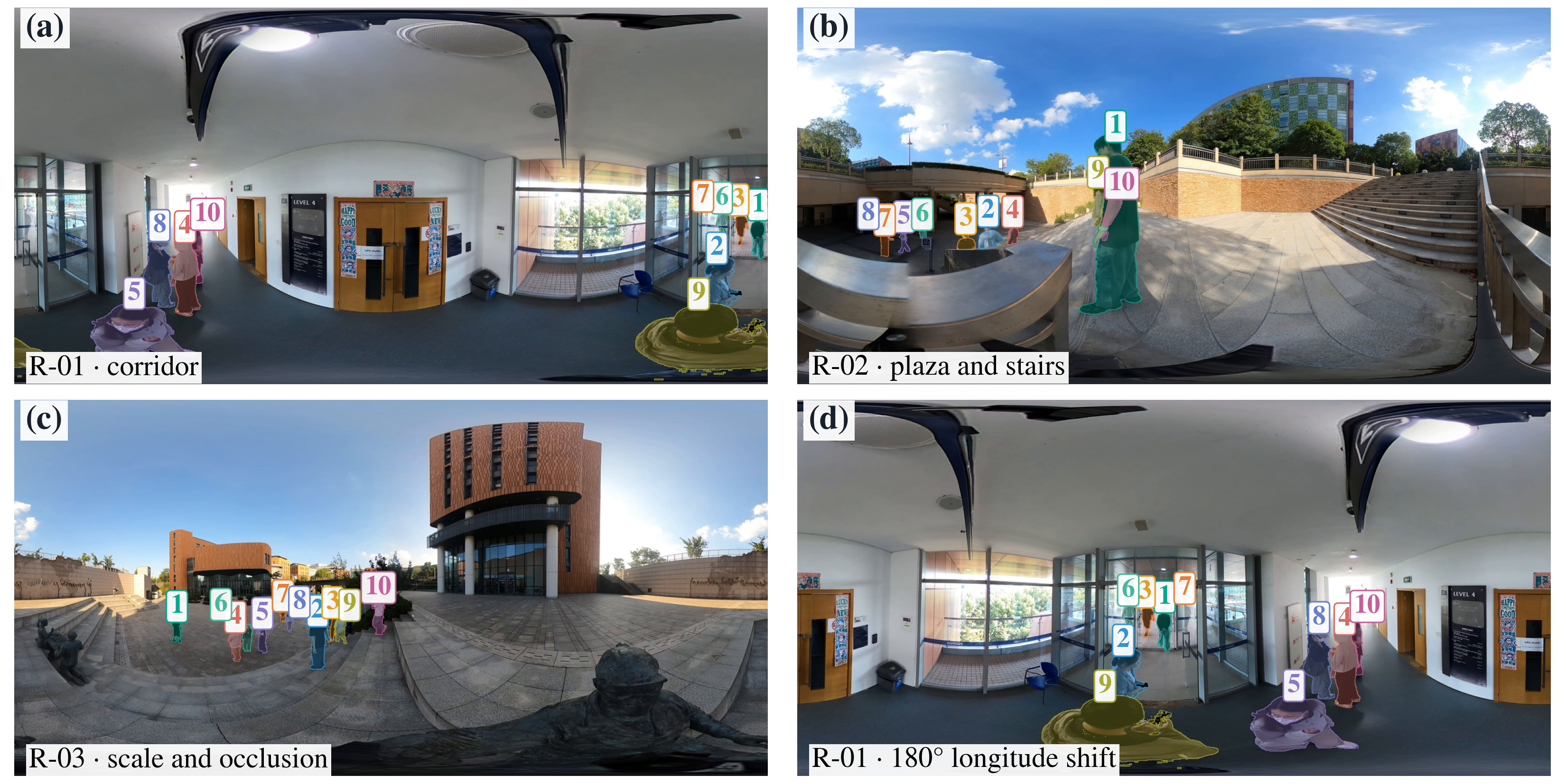}
\caption{\textbf{People move continuously across the panorama in \benchR{}.}
(a) R-01: a corridor with people crossing the seam and near the lower pole.
(b) R-02: a plaza on several levels with stairs. (c) R-03: large changes in
scale, and occlusion. (d) The frame of (a) after a $180^\circ$ horizontal
shift, which moves the seam into the middle of the image without changing the
scene. Colored overlays show the frozen Support-auto visible-support
predictions. Numbers are track identities, preserved across the shift.}
\label{fig:panoped_r_overview}
\end{figure}

\paragraph{Preparing the ERP images.}
The camera's two fisheye images were exported as gravity-aligned
$5376\times2688$ ERP video (HEVC, 29.97\,fps), with horizon leveling on and
direction lock and electronic stabilization off. We shifted the kept frames
horizontally at the source resolution, downsized them to $3840\times1920$ with
area resampling, and saved them as JPEG images at quality 95. The shift is
$x'=(x-s)\bmod 5376$. Every source offset $s$ is a multiple of seven, so the
corresponding shift at the released resolution, $5s/7$, is a whole number of
pixels. The shift only moves the starting point of longitude; it does not crop
the sphere and does not define the camera's absolute orientation. The
$180^\circ$ shift in Figure~\ref{fig:panoped_r_overview}d is for display only.
Only approved time windows are released: R-01 drops the first 840 source
frames, U01 ends before 210\,s, and R-03 before 145\,s.

\paragraph{Remaining camera motion.}
Although the cameras were fixed, we measure how much the image still moves
over time. Every five seconds, we estimate the image shift with phase
correlation, a standard method for finding the offset between two images, on a
$4\times8$ grid of tiles of $672\times672$ source pixels. Each tile is
mean-subtracted and tapered toward its borders with a Hann window; tiles whose
correlation response is below 0.1 are discarded, and we take the median shift
of the rest. We also measure shorter-term motion at the full frame rate in
separate ten-second intervals. For R-01, the five-second samples are
restricted to the released part of the video; the ten-second interval chosen
earlier still lies inside this part, but no longer at its center.
Table~\ref{tab:real_overview} reports the 95th percentile of the five-second
measurements and the horizontal and vertical ranges within the ten-second
intervals. These numbers describe image motion; they are not a calibration of
the camera's orientation.

\subsection{Annotation rules}
\label{app:real_annotation_contract}
\paragraph{Boxes, identities, and visibility.}
Table~\ref{tab:real_annotation_rules} summarizes the annotation rules. Each
box encloses the visible part of a person (a modal box) and does not guess
where the hidden body is. A person keeps the same identity through occlusion,
re-entry, and seam crossings whenever the video shows that it is the same
person. Regions excluded from evaluation are stored as polygons without person
identities.

The \texttt{outside} flag marks frames in which a person is absent or cannot
be reliably identified or located. A completely hidden person may keep a
position record with visibility 0.0 if the surrounding frames support that
position; such a record is not scored. Seeing a person before and after a gap
is not enough to justify interpolating their position through the gap.

Visibility describes how clearly a person can be recognized, not what fraction
of the body is visible. Its five levels range from clear to no evidence. A
person seen through glass starts at 0.8, and a person too far away to identify
is marked \texttt{outside}. Box coordinates are interpolated linearly between
keyframes, while visibility and occlusion labels stay the same until the next
annotated change. Annotators add keyframes wherever these rules would
otherwise misrepresent the video.

\begin{table}[H]
\centering\small
\caption{\textbf{Real annotation rules.} The rules treat three questions
separately: what is visible, which identity a person has, and whether their
location can be evaluated.}
\label{tab:real_annotation_rules}
\begin{tabularx}{\linewidth}{@{}>{\raggedright\arraybackslash}p{.13\linewidth}>{\raggedright\arraybackslash}X>{\raggedright\arraybackslash}X@{}}
\toprule
Property & Rule & Consequence \\
\midrule
Extent & Box around the visible body & Hidden full-body bounds are not invented. \\
Identity & Kept through occlusion, re-entry, and seam crossing when the video supports it & Disappearing alone does not reset an identity. \\
Visibility & Five levels of how clearly a person can be recognized; glass starts at 0.8 & Visibility is not the unoccluded fraction of the body. \\
Outside & No location when the person is absent or their location is uncertain & Positions are interpolated only where the video supports it. \\
Keyframes & Added where position, size, or state changes & Sudden events get denser annotation. \\
Seam & One identity; the box may wrap around the seam & Two image fragments remain one person. \\
Pole & Keep targets; check them in a tangent-plane view & Polar distortion is part of the benchmark. \\
Ignore & Height below 10\,px, visibility below 0.1, or a designated ignore region & Unreliable localization is not scored. \\
\bottomrule
\end{tabularx}
\end{table}

\subsection{Annotation workflow and identity checks}
\label{app:real_annotation_process}
Annotators reviewed and corrected the box tracks in CVAT. As documented, the
tracks were initialized automatically by linking YOLO11x detections with
ByteTrack~\citep{zhang2022bytetrack}, which provided candidate boxes and
temporary identities.\footnote{The retained import record for R-01 starts
after its ten tracks already existed, so it does not document how they were
initialized.} In manual passes, annotators corrected identity continuity,
fitted boxes to the visible body, and set keyframes, visibility, occlusion,
and \texttt{outside} states. Reviewers split identities that had been wrongly
merged, linked people who reappeared when the video supported the link, and
added keyframes where interpolation was inaccurate. Further checks went
through the occlusion intervals and followed each track over time and across
the ERP seam. Support-auto was derived afterwards from the final
human-reviewed boxes and did not replace the review of boxes or identities.

\subsection{Spherical Regions and MOT-Compatible Labels}
\label{app:real_truth}
For real data, we map pixels to directions with Equation~\ref{eq:erp_sphere},
which uses pixel centers. Its half-pixel terms separate integer pixel indices
from continuous coordinates at pixel edges, and our converters apply this
offset exactly once. In the real data, L0 stores the human boxes around the
visible body and the frames where states change. L1 stores the fitted regions
on the sphere, including their orientation and local bounds. L2 stores ERP
boxes and marks which horizontal extents were observed and which were
estimated across the seam (see below). L3 provides standard MOT rows.

\paragraph{Converting a box to a bounding field of view.}
The Box-fit layer describes each human box by a bounding
field of view (BFoV): a center
direction with an angular width and height. We sample at least 200 points
along each of the four box edges and map their pixel centers to directions on
the unit sphere. We then apply the smallest rotation $R_c$ that turns the
box's center direction $c$ to $+z$. In this rotated frame, we unwrap the
boundary longitudes so that they do not jump at $\pm180^\circ$, and read the
angular width and height from the range of longitudes and latitudes. Regions
taller than $90^\circ$ follow the extended BFoV convention of
\citet{huang2023vot}. The evaluator keeps the stored rotation and the possibly
asymmetric bounds, so a region is never silently treated as an unrotated,
symmetric box. To map a region back to ERP, the converter tests which pixel
rays fall inside its angular bounds. The converter can also return the stored
source box directly, but getting this box back does not mean that the box and
the fitted region have the same shape. Our tests check that saving and
reloading a region keeps all of its fields and, in the tested cases, that
rotating a region and the pixel rays together does not change which pixels lie
inside. Shifting an ERP box and fitting a new region is a different operation.
All methods are scored with the same evaluator.

\paragraph{Seam crossings and excluded regions.}
A track crosses the seam when its longitude, followed continuously over time,
passes the meridian where the image is cut. If a box comes within two pixels
of the left or right image border, we set the flag \texttt{seam\_truncated}.
L2 then keeps the fragment drawn by the annotator and, separately, an estimate
of the person's full horizontal extent taken from the angular width in nearby
frames where the person is not cut. This estimate is marked as such and never
replaces the human annotation. L3 scores the larger fragment and exports the
smaller one with confidence zero. Boxes shorter than ten pixels or with
visibility below 0.1 also get confidence zero. Ignore regions keep their
polygon form and, for MOT compatibility, are also exported as box rows with
confidence zero.

\begin{figure}[H]
\centering
\figfile{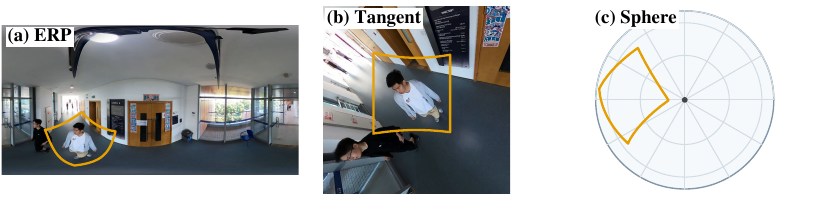}{3.6cm}{}
\caption{\textbf{Checking annotations near the lower pole.} The same released
R-01 target, identity 6 in frame 14, shown (a) in ERP, (b) in a gnomonic
(tangent-plane) view centered on the target, and (c) on the sphere in
orthographic projection. Orange traces the same oriented Box-fit boundary in
all three views. Real pixels are averaged in blocks before projection, while
the overlays are drawn sharp; reprojection does not change the annotation.}
\label{fig:real_polar_qc}
\end{figure}

\subsection{Quality control and release statistics}
\label{app:real_qc}
We check that L1, L2, and L3 contain exactly the same scoreable (frame,
identity) pairs, and that every scored L3 row refers to exactly one L1 record.
All three labeled sequences pass, with 47,439, 59,439, and 43,237 scoreable
boxes. Random rotations with a fixed seed check that region centers stay
aligned after projecting forward and back, and examples near the lower pole
are inspected in the tangent view of Figure~\ref{fig:real_polar_qc}. To look
for people the annotators may have missed, we take detections on every
fifteenth frame. We keep those with confidence at least 0.6 whose IoU with
every person and ignore annotation is below 0.3, and group them in time and
space. The 733 unmatched detections form 32 events for review: 28 in R-01 and
two in each of the other labeled sequences. These events are candidates for
review, not confirmed missing labels.

\begin{table}[H]
\caption{\textbf{Statistics of the real annotations.} ``Polar'' means that
the BFoV center lies below $-50^\circ$ latitude; ``Outside'' counts
\texttt{outside} intervals. Visibility counts include the unscored intervals
with visibility 0.0.}
\label{tab:real_statistics}
\centering\small
\setlength{\tabcolsep}{3.3pt}
\begin{tabular}{lrrrrrrrr}
\toprule
Seq. & IDs & Frames & Boxes & Scoreable & Seam & Polar & Keyframes & Outside \\
\midrule
R-01 & 10 & 5,486 & 47,543 & 47,439 & 10 & 3,151 & 5,983 & 36 \\
R-02 & 10 & 6,416 & 60,468 & 59,439 & 0 & 0 & 6,533 & 8 \\
R-03 & 10 & 4,345 & 43,450 & 43,237 & 0 & 0 & 4,735 & 0 \\
\midrule
Labeled & 30 & 16,247 & 151,461 & 150,115 & 10 & 3,151 & 17,251 & 44 \\
\bottomrule
\end{tabular}
\par\medskip
\begin{tabular}{lrrrrr}
\toprule
Seq. & Vis.\ 1.0 & 0.8 & 0.5 & 0.2 & 0.0 \\
\midrule
R-01 & 22,223 & 14,262 & 8,686 & 2,268 & 104 \\
R-02 & 31,344 & 16,456 & 9,328 & 2,311 & 1,029 \\
R-03 & 29,835 & 9,566 & 2,865 & 971 & 213 \\
\midrule
Total & 83,402 & 40,284 & 20,879 & 5,550 & 1,346 \\
\bottomrule
\end{tabular}
\end{table}

Table~\ref{tab:real_statistics} summarizes the released annotations; the 44
\texttt{outside} intervals cover 10,933 frames. Keyframes are placed where
they are needed rather than at fixed intervals, and most gaps between them
span six to ten frames. No labeled person box reaches the top 200 rows of the
image. The current labeled sequences need no ignore polygons for crowds or
camera mounts, although the annotation interface supports both.

\subsection{Automatic masks as a second real target (Support-auto)}
\label{app:support_auto}
\supportR{} v0.2 adds targets based on visible pixels to the same three real
videos. It leaves the human identities, boxes, visibility, \texttt{outside}
states, and Box-fit records unchanged. For each observation $i$, a frozen
segmentation model prompted with the human box predicts a mask in a local view
around the person. The mask is mapped back to ERP and converted with the same
operator as in S:
\begin{equation}
 \widetilde M_i=A_\psi(I_i,B_i^{\mathrm{ann}}),\qquad
 \widetilde Y_i=F_S(\widetilde M_i).
\label{eq:auto_target}
\end{equation}
$F_S$ computes the area-weighted center on the sphere and the angular spans in
a local frame aligned with gravity. The main layer uses SAM2.1 Hiera-large
(224.45M parameters); SAM-HQ ViT-B provides a second target layer for
sensitivity analysis \citep{ravi2025sam2,ke2023samhq}. Human boxes are used
only to build these annotations and are never given to a tracker. The masks
are automatic estimates, not independent human segmentations. We fixed this
second protocol after exploratory comparisons of targets, so it is not a new
blind dataset.

\begin{table}[H]
\centering\small
\caption{\textbf{Coverage of the automatic masks.} Unknown: observations that
were scoreable in Box-fit but have no valid mask. Filters used to select training
data never remove evaluation targets. Common valid: observations with a valid
mask from both SAM2 and SAM-HQ.}
\label{tab:support_coverage}
\setlength{\tabcolsep}{5pt}
\begin{tabular}{@{}lrrrrr@{}}
\toprule
Sequence & Frames & Box-fit scoreable & SAM2 valid & Unknown & Common valid \\
\midrule
R-01 & 5,486 & 47,439 & 47,257 & 182 & 47,184 \\
R-02 & 6,416 & 59,439 & 58,932 & 507 & 58,886 \\
R-03 & 4,345 & 43,237 & 43,013 & 224 & 42,732 \\
\midrule
Total & 16,247 & 150,115 & 149,202 & 913 & 148,802 \\
\bottomrule
\end{tabular}
\end{table}

\paragraph{Scoring observations without a valid mask.}
The 1,346 observations that were already unscored in Box-fit stay unscored.
Observations with a valid mask form $\mathcal V_t$, and the 913 further
observations without a valid mask form $\mathcal U_t$. A prediction that
matches an observation in $\mathcal U_t$ cannot be scored, and counting it as
a false positive would penalize the tracker for a gap in our labels. Before
native scoring, we match each tracker's original prediction boxes to the human
boxes in $\mathcal V_t\cup\mathcal U_t$ at periodic IoU $\ge0.5$, maximizing
first the number of matches and then the total IoU. A prediction matched to an
observation in $\mathcal U_t$ is excluded from scoring, and each such
observation excludes at most one prediction. All other predictions are kept,
and each readout is then scored independently against $\mathcal V_t$. This
step only removes predictions; it does not decide which predictions count as
correct, and one observation cannot excuse several duplicates. Because the
matching uses the tracker's boxes, all readouts of the same tracker exclude
the same predictions. All 16,247 frames and the original identities are kept,
including gaps without a valid mask, and metrics are computed from counts
summed over whole sequences. Appendix~\ref{app:support_results} also reports
two alternative policies: one that excludes no additional predictions, and one
restricted to observations valid for both SAM2 and SAM-HQ.

\paragraph{Release checks.}
Version 0.2 recomputes all targets from the unchanged v0.1 masks, stored as
run-length encodings, with the same frozen $F_S$; the largest reported
difference is about $3\times10^{-13}$ degrees, that is, numerical round-off.
As in v0.1, the manifests lack generation-time hashes for six external
modules. The quality filter used to select training data keeps 126,783
observations; it is never applied during evaluation. Under the common-valid
policy, 1,313 observations have no valid mask, and the longest gaps without
one last 21, 275, and 38 frames in the three sequences. Coverage and agreement
between the two mask generators show that the masks are available and stable;
they do not show that the masks are accurate.

\subsection{Scope and limitations}
\label{app:real_limitations}
\benchR{} focuses on dense, continuous identity annotation in five views with
ten participants. Its three labeled sequences allow LOSO evaluation, but they
cover few scenes and participants, and the fixed cameras complement, but do
not reproduce, the moving-platform conditions of \benchS{}. Camera height and
metric scene geometry are not available. The Box-fit regions come from
boxes and can be loose near the lower pole; an extent that was estimated over
time across the seam is marked as such in L2. Support-auto adds automatic
masks, not independent human masks; its coverage and its sensitivity to the
mask generator are reported in Appendix~\ref{app:support_auto}. Finally, the
zero of the yaw angle, which determines where the seam lies, is an arbitrary
choice of coordinates.

\section{Release, Licensing, and Reproducibility}
\label{app:release}
The release contains rendered ERP images, numerical annotations, and benchmark
code. It does not contain the source meshes, textures, or engine packages, and
they cannot be recovered from it. The recorded asset licenses cover the
RocketBox people and motions, CC0 environment assets, the BSD-3-Clause
quadruped, and the CC-BY-4.0 drone. Facade assets from the marketplace appear
only in rendered form, following an item-by-item release check. Attributions,
notices, and the exact boundaries of what is released are listed in the
datasheet instead of being repeated here.

The archive needed for training contains RGB images and tracking labels;
instance and depth rasters come as optional archives. Compressed, the training
archive takes \PanoPedSTrainingArchiveGiB{}\,GiB and the complete release
\PanoPedSCompleteArchiveGiB{}\,GiB. The R release follows its research-only
license and the participants' consent conditions. For reproducibility, we
release sequence configurations, versioned target operators, split
definitions, baseline settings, hashes of models and predictions, and the
evaluator. Appendix~\ref{app:dataset} summarizes which assets are reused
across splits; holding out sequences does not give each split its own scene
family or avatar pool.

\section{Supplementary Experiments}
\label{app:experiments}
\subsection{Experiment Setup and Evaluation Checks}
\method{} changes only the final region of each output; the detector query
index, confidence, and identity are kept. The main comparisons are the
geometric readout, the ERP-supervised query head, the geometry-only head, and
\method{}. The capacity-matched geometry head has about the same number of
parameters as \method{} but a different hidden width and network form. We also
record the order in which decisions were made. Both \method{} heads were fixed
before we evaluated on V-audit. The ERP and geometry-only controls were added
after this first result, and the capacity-matched control later still, without
retuning the \method{} heads. The rule for choosing among e3, e4, and e5 was
fixed before the e5 scores were seen, and the test weights, code, and scoring
settings were fixed before test evaluation. The two head seeds share one
detector training run; the validation splits and the bootstrap resamples do
not add independent detector trainings.

\paragraph{What is evaluated.}
HOTA averages over all 19 native-IoU thresholds. Within each comparison, all
methods use the same confidence policy, scoreable targets, and ignore
handling. For raw detector AP, we keep the 100 highest-scoring boxes per image
with a stable sort, apply neither non-maximum suppression nor a confidence
threshold, and count every finite box that matches no target as a false
positive. Tracker outputs are filtered differently, so we never report AP on
tracker outputs in place of raw detector AP. The geometric errors in
Section~\ref{sec:analysis} are computed on a fixed set of matches chosen in
advance, and we report them separately from the sequence-level tracking
metrics.

\paragraph{Detector accuracy versus tracking.}
The e3 and e4 checkpoints reach V-dev AP50/AP75 of 84.91/53.76 and
84.25/55.66, yet their tracking scores fall from 49.06/56.58 to 45.74/53.14
HOTA/IDF1. D-FINE reaches AP75 of 59.29 and 60.78 with its two seeds, yet its
OC-SORT systems score only 32.51 and 34.71 V-dev HOTA. Detector AP is
therefore an incomplete guide to tracking quality on the sphere, because
tracking also depends on which detections are passed on as candidates and on
decisions made over time.

\paragraph{Scope of the study.}
The evaluated method is only the query-conditioned localization head. We also
explored variants for identity assignment, readouts from masks, and targets
split into separate parts; they are recorded in the development log and are
not part of the evaluated system.

\subsection{Uncertainty estimated by resampling sequences}
\begin{figure}[H]
\centering
\figfile{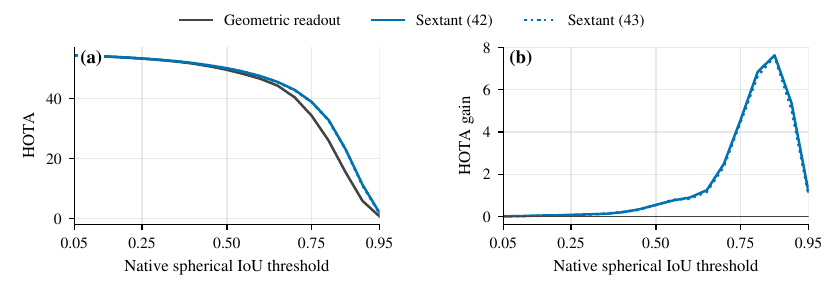}{5.0cm}{}
\caption{\textbf{Localization tolerance on V-audit.} (a) HOTA at each of the
19 native-IoU thresholds. (b) Gain over the geometric readout. Both heads use
the baseline's identities, so the change in gain across thresholds comes only
from the output regions.}
\label{fig:localization_thresholds}
\end{figure}

We estimate uncertainty by resampling whole sequences, because frames within a
sequence are strongly correlated and cannot be treated as independent samples.
V-audit has four sequences, so we enumerate all $4^4=256$ resamples, each
drawing four sequences with replacement, and recompute the metrics from the
summed raw counts of each resample. The 2.5--97.5 percentile ranges of the
HOTA gain are $[0.87,2.31]$ and $[0.72,2.25]$ for the two seeds. On test, we
draw 2,000 paired resamples of whole sequences with a fixed random seed, which
gives the ranges reported in Section~\ref{sec:main_results}.
Figure~\ref{fig:localization_thresholds} repeats the threshold analysis on
V-audit, which was held back until the heads were fixed. Every test sequence
improves over the geometric readout. HAT, however, is much stronger than the
MOTIP-based systems on the drone sequence SE-030, so a better overall ranking
does not mean better results in every scene.

\begin{table}[H]
\centering\small
\caption{\textbf{All readouts on all splits of PanoPed-S.} HOTA and IDF1 on the three
splits for both head seeds; Table~\ref{tab:controls} shows the HOTA columns.
All rows use the same frozen MOTIP detector and the same identities.}
\label{tab:controls_full}
\setlength{\tabcolsep}{3pt}
\begin{tabular}{@{}lcrrrrrr@{}}
\toprule
& & \multicolumn{2}{c}{V-dev (7,200 fr.)} & \multicolumn{2}{c}{V-audit (7,200 fr.)} & \multicolumn{2}{c}{Test (14,400 fr.)} \\
\cmidrule(lr){3-4}\cmidrule(lr){5-6}\cmidrule(l){7-8}
Readout & Seed & HOTA & IDF1 & HOTA & IDF1 & HOTA & IDF1 \\
\midrule
Geometric (area centroid) & -- & 49.06 & 56.58 & 41.41 & 47.09 & 47.30 & 55.70 \\
ERP-supervised query head & 42 & 49.09 & 56.55 & 41.39 & 47.04 & 47.24 & 55.65 \\
ERP-supervised query head & 43 & 49.12 & 56.56 & 41.41 & 47.05 & 47.26 & 55.64 \\
Geometry-only head & 42 & 49.27 & 56.62 & 41.77 & 47.14 & 47.61 & 55.75 \\
Geometry-only head & 43 & 49.21 & 56.63 & 41.81 & 47.17 & 47.50 & 55.78 \\
Capacity-matched geometry head & 42 & 49.09 & 56.65 & 41.63 & 47.18 & 47.52 & 55.80 \\
Capacity-matched geometry head & 43 & 49.32 & 56.66 & 41.83 & 47.19 & 47.64 & 55.79 \\
\textbf{\method{}} & 42 & \underline{51.03} & \textbf{56.81} & \textbf{43.12} & \textbf{47.38} & \textbf{49.49} & \textbf{56.06} \\
\textbf{\method{}} & 43 & \textbf{51.04} & \underline{56.80} & \underline{43.06} & \underline{47.37} & \underline{49.47} & \underline{56.05} \\
\bottomrule
\end{tabular}
\end{table}

\begin{table}[H]
\centering\small
\caption{\textbf{HOTA components on PanoPed-S test.} Changing the output regions changes
which predictions the evaluator matches, and therefore both DetA and AssA,
even though the tracker's identity decisions are unchanged.}
\begin{tabular}{@{}lrrr@{}}\toprule
Readout & DetA & AssA & LocA \\ \midrule
Geometric readout & 62.31 & 36.05 & 82.54 \\
\method{} 42 & 64.90 & 37.94 & 84.67 \\
\method{} 43 & 64.85 & 37.94 & 84.64 \\
\bottomrule\end{tabular}
\end{table}

\subsection{Per-sequence test results}
\label{app:per_sequence}
Table~\ref{tab:per_sequence} lists HOTA and IDF1 for the geometric readout and
both \method{} heads on each test sequence. Identity metrics change only
because the evaluator matches predictions differently. The number of ID
switches moves from 2,552 to 2,584 and 2,568 on V-audit, and from 7,514 to
7,608 and 7,627 on test; all original predictions and scores are still passed
to the evaluator.

\begin{table}[H]
\centering\small
\caption{\textbf{Results on each PanoPed-S test sequence.} HOTA and IDF1 for each of the
eight test sequences; \method{} seeds 42\,/\,43. The totals match
Table~\ref{tab:final_test}; they are computed from summed raw counts, not by
averaging the per-sequence scores.}
\label{tab:per_sequence}
\setlength{\tabcolsep}{4pt}
\begin{tabular}{@{}llrrrrrr@{}}
\toprule
& & \multicolumn{2}{c}{Geometric} & \multicolumn{2}{c}{\method{} 42\,/\,43} & \multicolumn{2}{c}{Gain 42\,/\,43} \\
\cmidrule(lr){3-4}\cmidrule(lr){5-6}\cmidrule(l){7-8}
Sequence & Platform & HOTA & IDF1 & HOTA & IDF1 & HOTA & IDF1 \\
\midrule
SS-027 & Fixed & 58.08 & 71.85 & 60.51\,/\,60.83 & 72.29\,/\,72.21 & +2.43\,/\,+2.75 & +0.44\,/\,+0.36 \\
SS-028 & Fixed & 41.23 & 46.95 & 43.38\,/\,43.35 & 47.42\,/\,47.44 & +2.15\,/\,+2.12 & +0.48\,/\,+0.50 \\
SS-029 & Fixed & 61.57 & 75.22 & 65.27\,/\,65.12 & 75.96\,/\,75.94 & +3.69\,/\,+3.54 & +0.74\,/\,+0.71 \\
SS-030 & Fixed & 38.12 & 44.01 & 39.94\,/\,39.81 & 44.32\,/\,44.27 & +1.82\,/\,+1.69 & +0.31\,/\,+0.26 \\
SE-017 & Quadruped & 31.12 & 39.47 & 31.89\,/\,31.78 & 39.51\,/\,39.56 & +0.77\,/\,+0.66 & +0.04\,/\,+0.09 \\
SE-018 & Quadruped & 32.24 & 38.25 & 33.38\,/\,33.33 & 38.41\,/\,38.41 & +1.13\,/\,+1.09 & +0.16\,/\,+0.16 \\
SE-029 & Drone & 55.01 & 64.75 & 57.46\,/\,57.40 & 65.00\,/\,65.01 & +2.45\,/\,+2.39 & +0.25\,/\,+0.26 \\
SE-030 & Drone & 49.37 & 59.09 & 51.25\,/\,51.23 & 59.35\,/\,59.34 & +1.88\,/\,+1.86 & +0.26\,/\,+0.25 \\
\midrule
All (14,400 frames) & & 47.30 & 55.70 & 49.49\,/\,49.47 & 56.06\,/\,56.05 & +2.19\,/\,+2.17 & +0.36\,/\,+0.35 \\
\bottomrule
\end{tabular}
\end{table}

\subsection{Inference cost}
\label{app:runtime}
\begin{table}[H]
\centering\small
\caption{\textbf{Throughput of the complete pipeline in frames per second on PanoPed-S.}
Two fixed training sequences of 1,800 frames each; the first 30 frames are
warm-up. On each machine, the two runs execute at the same time on separate
GPUs. The timing includes the final localization step but not metric
evaluation.}
\label{tab:runtime}
\begin{tabular}{llrrr}
\toprule
GPU & Sequence & Geometric & \method{} 42 & \method{} 43 \\
\midrule
RTX 5090 D v2 & SE-001 & 11.65 & 11.55 & 11.57 \\
RTX 5090 D v2 & SS-001 & 11.05 & 11.14 & 11.15 \\
RTX 4090 D & SE-001 & 6.03 & 6.17 & 6.18 \\
RTX 4090 D & SS-001 & 6.09 & 6.17 & 6.15 \\
\bottomrule
\end{tabular}
\end{table}

Table~\ref{tab:runtime} reports throughput on two training sequences chosen in
advance, excluding the first 30 frames as warm-up and running one sequence per
GPU at the same time. Input images are $2048\times1024$, and the detector
resizes them within its usual limits of 960 pixels on the short side and 1,920
on the long side. The timed pipeline covers decoding, resizing, copying to the
GPU, detection, identity tracking, the final localization step, and collecting
the outputs on the CPU; it excludes writing to disk and metric evaluation.
Within each run, the detector and identity outputs have identical hashes for
all readouts. The CUDA allocator uses at most 950--952 MiB, not counting
driver, library, and host memory. The two runs on a machine share storage and
CPU, so small differences in timing do not show a speedup. The two machines
also use different software environments, so the numbers cannot isolate the
effect of the GPU model. Head training, detector training, feature extraction,
development experiments, and evaluation are separate costs; the cluster
records do not give an exact total of GPU hours for the whole project.

\subsection{Training budgets and tracker settings of the baselines}
\label{app:budgets}
The selected synthetic MOTIP starts from COCO weights and is trained for 20
detection epochs and then three joint tracking epochs; the checkpoints after
the fourth and fifth joint epochs are the only alternatives we compare. Every
head follows the recipe of Section~\ref{sec:training}, and using another
detector checkpoint requires extracting its features and training new heads.
ByteTrack and OC-SORT use the same e3 detections without further training.
D-FINE-L \citep{peng2025dfine} starts from a COCO checkpoint and is trained
for 12,000 updates at batch size four, that is, 48,000 sampled images per
seed. Its training set leaves out three sequences that we used for internal
development, whereas the MOTIP head uses all 44 training sequences. This
reproduction under a limited budget keeps horizontal flips and a constant
fine-tuning learning rate. OmniTrack uses its own detector and its final
checkpoint after 565,700 updates, including the documented reset during
recovery. HAT uses frozen MASA-R50 features without further appearance
training on \bench{}. These differences are part of the system comparison, so
not every difference in score can be attributed to the architecture alone.

The fixed BoxMOT implementations use maximum age 30, minimum hits three, and
IoU threshold 0.3. ByteTrack uses minimum confidence 0.05, track threshold
0.3, match threshold 0.9, and a buffer sized for the nominal 30 fps. OC-SORT
uses minimum confidence 0.1, detection threshold 0.3, delta time three, and
inertia 0.1, with its ByteTrack option turned off. Neither tracker receives a
ReID, seam, or camera-motion module. The baseline registry records the
remaining settings and the hashes of the source code and weights.

\paragraph{OmniTrack on real data.}
All real folds are rescored with the same published box-fit metric for R. Fold
1 still uses an earlier version of the code that prepares the training data
and target coordinates; its archived predictions left out 649 Kalman boxes
with non-positive width. The rescoring keeps every archived valid row and adds
no exclusions. In fold 2, boxes with positive width that span a full panorama
period are kept by the common R fit. We report this implementation difference
openly rather than treating the training code as identical across folds.

\subsection{Evaluation with Box-fit targets}
\label{app:transfer}
Box-fit scores oriented regions fitted to the annotated periodic boxes,
whereas \method{} predicts mask-based centers and spans in a gravity-aligned
frame. Its four outputs do not implement the Box-fit operator, so
Table~\ref{tab:real_protocols} reports only the box-based baselines that are
compatible with Box-fit. Scoring \method{}'s predictions directly against
Box-fit did not give an improvement; these cross-target runs are kept in the
evaluation archive and are not presented as another way to use \method{}.

\begin{table}[H]
\centering\small
\caption{\textbf{Evaluation with Box-fit targets.} Each
LOSO fold trains on two R sequences and tests on the third; the total covers
16,247 held-out frames, each scored once with the R target operator. The
sim-to-real (S$\to$R) columns use no real labels for training. All systems use
the box-based R converter. Bold and underline mark the best and second-best
value in each column.}
\label{tab:real_protocols}
\setlength{\tabcolsep}{4.5pt}
\begin{tabular}{@{}llrrrr@{}}
\toprule
& & \multicolumn{2}{c}{Real-only LOSO} & \multicolumn{2}{c}{Sim-to-real S$\to$R} \\
\cmidrule(lr){3-4}\cmidrule(l){5-6}
Method & Venue & HOTA & IDF1 & HOTA & IDF1 \\
\midrule
OmniTrack & CVPR'25 & 12.06 & 9.63 & 17.86 & 15.26 \\
ByteTrack & ECCV'22 & 28.65 & 24.54 & 28.21 & 23.71 \\
OC-SORT & CVPR'23 & 24.89 & 20.96 & 30.73 & 26.69 \\
MOTIP & CVPR'25 & 24.04 & 20.59 & 34.60 & \underline{32.70} \\
HAT, history off & control & \underline{28.86} & \underline{25.87} & \textbf{35.75} & 32.27 \\
HAT + MASA & ECCV'26 & \textbf{29.97} & \textbf{27.80} & \underline{34.86} & \textbf{32.75} \\
\bottomrule
\end{tabular}
\end{table}

\subsection{Qualitative examples selected by a fixed rule}
\label{app:qualitative}
\begin{figure}[!htbp]
\centering
\includegraphics[width=\linewidth,height=.64\textheight,keepaspectratio]{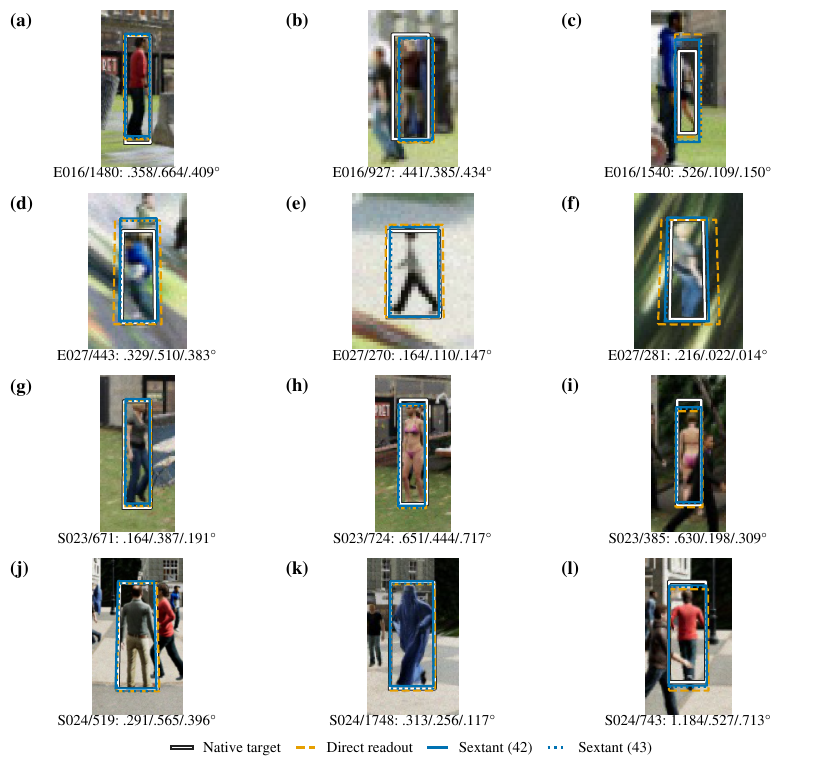}
\caption{\textbf{Twelve localization examples selected by a fixed rule.} Rows
show SE-016, SE-027, SS-023, and SS-024; columns show the 10th, 50th, and 90th
percentiles of the improvement in center error. White: native target; orange
dashed: direct rectangle readout; blue solid and dotted: \method{} seeds 42
and 43. Each label gives the sequence and frame first, then the center errors
in degrees; S/E abbreviate SS/SE. The reference is the direct rectangle
readout, not the area-centroid conversion. The crops keep the original pixels
and wrap around the seam where needed; enlarging them adds no detail. A better
center alone does not imply a better extent or identity.}
\label{fig:qualitative_localization}
\end{figure}

The qualitative examples are chosen by a fixed rule from the 117,875 V-dev
pairs matched by the original direct rectangle readout at native IoU $\ge0.5$.
Within each sequence, we rank the pairs by the center error of the direct
readout minus the mean error of the two \method{} heads. We then select the
pairs at positions $\lfloor q(N-1)\rfloor$ for $q\in\{0.1,0.5,0.9\}$ in
ascending order, breaking ties by frame and query index. The selection is made
before any image is decoded. Figure~\ref{fig:qualitative_localization} shows
the result. The examples therefore include cases where \method{} is worse and
where the two seeds disagree, as well as large gains. They illustrate the
method after development was finished; they are neither an ensemble nor an
independent test sample.

\subsection{Full results on real data}
\label{app:support_results}
\paragraph{Training conditions and controls.}
In the sim-to-real setting, the original S e3 detector and the S-trained heads
stay frozen. For real-only LOSO, each fold has its own detector trained on the
other two R sequences, and a head for the support targets is trained on the
same two sequences. Both kinds of head follow Section~\ref{sec:training}:
25,000 updates, batch size 256, seeds 42 and 43. These heads predict the
support targets directly; they are not models with an extra box or support
output. For each seed, the three held-out prediction sets are combined without
choosing a different setting per fold. HAT keeps its external MASA
pretraining, and the R-trained heads also use the offline SAM2 support labels.

The area-centroid readout $U$ uses no network. The geometry head has 34,687
parameters and \method{} 34,692. For the separate comparison with an extra
image encoder, a frozen SAM2.1 Hiera-large (224.45M parameters) segments the
current \emph{predicted} box, and $F_S$ converts the mask; it never sees an
annotated test box. When SAM fails, or when an output has no detector query in
the current frame, the output keeps its $U$ region, so output counts,
identities, and confidences do not change. No readout feeds back into the
track history. The final banks of training matches for the R heads are built
with $U$. A first implementation built these matches from rectangle centers;
we corrected it by rebuilding all six banks and twelve heads, and all reported
results use the corrected runs.

\begin{table}[H]
\centering\small
\caption{\textbf{Complete comparison on real data under two training
settings.} All 16,247 frames, with the same Support-auto targets and the same
handling of observations without a valid mask. Sim-to-real uses the frozen S
models; real-only uses three separately trained LOSO models. SAM2.1 adds a
224.45M-parameter image encoder at inference. Bold marks the best score among
methods without this extra encoder; results with the extra encoder are listed
separately. The released machine-readable results contain further metric
components.}
\label{tab:support_source_full}
\label{tab:support_ronly_full}
\setlength{\tabcolsep}{4.5pt}
\begin{tabular}{@{}llrrrr@{}}
\toprule
 & & \multicolumn{2}{c}{Sim-to-real} & \multicolumn{2}{c}{Real-only LOSO} \\
\cmidrule(lr){3-4}\cmidrule(l){5-6}
Tracker & Readout & HOTA & IDF1 & HOTA & IDF1 \\
\midrule
MOTIP & $U$ & 31.6039 & 31.9601 & 21.2115 & 20.0724 \\
HAT-off & $U$ & 32.5944 & 31.8041 & 25.4321 & 24.8853 \\
HAT & $U$ & 31.7109 & 31.9482 & 26.4901 & 27.0553 \\
ByteTrack & $U$ & 25.5573 & 23.2654 & 25.3455 & 24.1897 \\
OC-SORT & $U$ & 28.0455 & 26.4405 & 22.1666 & 20.6875 \\
OmniTrack & $U$ & 15.9270 & 14.5172 & 10.5636 & 8.7375 \\
\midrule
MOTIP & Geometry 42 & 32.1869 & 32.0824 & 21.6973 & 20.1544 \\
 & Geometry 43 & 32.3209 & 32.1767 & 21.6745 & 20.1702 \\
\ours MOTIP & \ours\method{} 42 & \ours 32.7385 & \ours 32.2330 & \ours 21.0724 & \ours 19.7740 \\
\ours & \ours\method{} 43 & \ours 32.5614 & \ours 32.2349 & \ours 21.0994 & \ours 19.8477 \\
HAT & Geometry 42 & 32.3016 & 32.0484 & \textbf{27.0447} & 27.1579 \\
 & Geometry 43 & 32.4533 & 32.2153 & 27.0161 & \textbf{27.1843} \\
\ours HAT & \ours\method{} 42 & \ours\textbf{32.8500} & \ours 32.2759 & \ours 26.2266 & \ours 26.4108 \\
\ours & \ours\method{} 43 & \ours 32.6788 & \ours\textbf{32.2793} & \ours 26.2805 & \ours 26.4511 \\
\midrule
\multicolumn{6}{@{}l}{\emph{Additional segmentation encoder at inference}} \\
MOTIP & SAM2.1 & 38.0707 & 32.6073 & 26.1117 & 20.4176 \\
HAT & SAM2.1 & 38.2402 & 32.6837 & 32.4747 & 27.4056 \\
\bottomrule
\end{tabular}
\end{table}

\paragraph{Transfer versus training on little real data.}
The synthetic-trained heads improve both metrics with both trackers and beat
their geometry-only controls. HAT+\method{} leads the sim-to-real systems that
add no localization encoder, and HAT+SAM2.1 scores highest overall. Training
on real data alone does not reproduce the gain of the query head: for HAT, the
geometry head improves HOTA by 0.53--0.55, whereas \method{} lowers it by
0.21--0.26. With only two training sequences per fold, overfitting to the few
scenes is a plausible explanation, but we have not isolated its cause. The
quality of the detector queries and the optimization also differ from
synthetic training. We therefore use R mainly to assess transfer and keep LOSO
as a small-data reference, without treating the two training settings as
interchangeable.

\subsection{Sensitivity to automatic masks and evaluation rules}
\begin{table}[H]
\centering\small
\caption{\textbf{HAT HOTA for different targets and different handling of
observations without a valid mask.} \method{} and Geometry are averages over
the two seeds. The two common-valid rows use the same set of observations,
which can differ from the set in the main table.}
\label{tab:support_sensitivity}
\setlength{\tabcolsep}{3.8pt}
\begin{tabular}{@{}llrrrr@{}}
\toprule
Training & Target / handling & $U$ & Sextant & Geometry & SAM2.1 \\
\midrule
Sim-to-real & Main & 31.7109 & 32.7644 & 32.3774 & 38.2402 \\
 & No extra exemptions & 31.6186 & 32.6685 & 32.2829 & 38.1255 \\
 & SAM2, common valid & 31.7600 & 32.8157 & 32.4274 & 38.3036 \\
 & HQ, common valid & 31.2109 & 32.2591 & 31.9964 & 36.0244 \\
\midrule
Real-only & Main & 26.4901 & 26.2535 & 27.0304 & 32.4747 \\
 & No extra exemptions & 26.4181 & 26.1826 & 26.9574 & 32.3849 \\
 & SAM2, common valid & 26.5269 & 26.2914 & 27.0681 & 32.5233 \\
 & HQ, common valid & 26.1052 & 25.8485 & 26.7028 & 30.8292 \\
\bottomrule
\end{tabular}
\end{table}
The sim-to-real gain remains when the targets are generated with SAM-HQ
instead of SAM2, and when no additional predictions are excluded for
observations without a valid mask. The losses from real-only training also
remain. These tests show how much the results depend on how the labels are
generated and scored; they do not measure the accuracy of the real masks.

\begin{table}[H]
\centering\small
\caption{\textbf{Check for bias toward the mask generator on PanoPed-S.} Native IoU on the
same 899 query/person pairs from 48 quality-check frames of S TRAIN (not S
TEST), scored against rendered masks and against SAM2 masks.}
\label{tab:sam_affinity}
\begin{tabular}{@{}lrr@{}}
\toprule
Readout & Rendered-mask target & SAM2 automatic target \\
\midrule
$U$ & 0.819213 & 0.800618 \\
\method{} 42 & 0.851703 & 0.836522 \\
\method{} 43 & 0.850484 & 0.835707 \\
Predicted-box SAM2.1 & 0.872066 & 0.947758 \\
\bottomrule
\end{tabular}
\end{table}
On these identical objects, the gain of direct SAM over $U$ is about 0.09429
IoU larger when the targets come from SAM itself than when they come from
rendered masks. Its strong Support-auto score should therefore be read
together with this bias toward its own masks and with the cost of its extra
encoder. Agreement between two automatic mask generators does not replace
independent human masks.

\subsection{Real per-sequence results and readout cost}
\begin{figure}[!htbp]
\centering
\includegraphics[width=\linewidth]{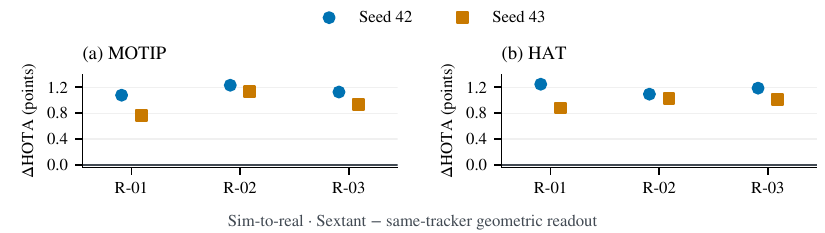}
\caption{\textbf{Sim-to-real gains on each complete real sequence.} Both seeds
of the synthetic-trained head improve HOTA with MOTIP and with HAT. Evaluation
uses Support-auto; no real labels are used for training.}
\label{fig:support_transfer}
\end{figure}
\begin{table}[H]
\centering\small
\caption{\textbf{Gains on each real sequence over the same tracker's $U$.}
Entries are $\Delta$HOTA\,/\,$\Delta$IDF1; no seed or readout is chosen per
fold.}
\label{tab:support_sequence}
\setlength{\tabcolsep}{3.8pt}
\begin{tabular}{@{}llcrrr@{}}
\toprule
Training & Tracker & Seed & R-01 & R-02 & R-03 \\
\midrule
Sim-to-real & MOTIP & 42 & $+1.0783/+0.1703$ & $+1.2323/+0.3652$ & $+1.1270/+0.2466$ \\
 & MOTIP & 43 & $+0.7596/+0.1632$ & $+1.1283/+0.4406$ & $+0.9315/+0.1658$ \\
 & HAT & 42 & $+1.2479/+0.2941$ & $+1.0940/+0.4030$ & $+1.1873/+0.2650$ \\
 & HAT & 43 & $+0.8803/+0.2893$ & $+1.0281/+0.4642$ & $+1.0063/+0.2049$ \\
\midrule
Real-only & MOTIP & 42 & $-0.7611/-0.8031$ & $-0.1570/-0.0248$ & $+0.3093/-0.1757$ \\
 & MOTIP & 43 & $-0.6837/-0.4837$ & $-0.0856/-0.0460$ & $+0.2392/-0.2078$ \\
 & HAT & 42 & $-1.0986/-2.0786$ & $-0.3911/-0.2383$ & $+0.4154/+0.1582$ \\
 & HAT & 43 & $-0.8795/-1.3362$ & $-0.1255/-0.1902$ & $+0.1253/-0.4463$ \\
\bottomrule
\end{tabular}
\end{table}

A separate microbenchmark on an H200 GPU measured 0.71--0.74\,ms for the query
head and decoding at batches of 8, 32, and 128 observations (five repetitions
of 1,000 iterations). It uses the same architecture with the weights from
before the matching correction described above. It excludes detector and
tracker work, data transfers, and metric scoring, so it is not an end-to-end
throughput on R. Table~\ref{tab:runtime} gives the complete-pipeline timing on
S. In the sim-to-real setting, SAM processed 37,669, 64,465, and 56,292
prompts in R-01, R-02, and R-03 and fell back to $U$ 0, 476, and 39 times; in
the real-only setting, it processed 37,149, 54,790, and 51,563 prompts with 0,
453, and 5,401 fallbacks. Every fallback keeps the output and uses $U$. The
full 32-row result tables, metric components, and results at all 19 thresholds
are kept in the evaluation archive. Neither the SAM encoder nor the offline
mask generation is counted in the 34,692 parameters of \method{}.

\end{document}